\documentclass{article}

\usepackage{microtype}
\usepackage{graphicx}
\usepackage{subfigure}
\usepackage{booktabs} 

\usepackage{hyperref}

\usepackage[accepted]{icml2024}

\usepackage{amsmath}
\usepackage{amssymb}
\usepackage{mathtools}
\usepackage{amsthm}

\usepackage[capitalize,noabbrev]{cleveref}

\usepackage{algorithm,algpseudocode}

\theoremstyle{plain}

\theoremstyle{definition}

\theoremstyle{remark}

\usepackage[textsize=tiny]{todonotes}

\usepackage[compact]{titlesec}
\usepackage{sidecap}
\titlespacing*{\section}{0pt}{*1.0}{*0.8}
\titlespacing*{\subsection}{0pt}{*0.5}{*0.3}
\titlespacing*{\subsubsection}{0pt}{*0.3}{*0.3}

\allowdisplaybreaks

\usepackage{hyperref}       
\usepackage{url}            
\usepackage{booktabs}       
\usepackage{amsfonts}       
\usepackage{nicefrac}       
\usepackage{microtype}      
\usepackage{xspace}
\usepackage{paralist}

\usepackage{wrapfig}
\usepackage{amsmath}
\usepackage{amssymb}
\usepackage{amsthm}
\usepackage{url}
\usepackage{paralist}
\usepackage{latexsym}
\usepackage{arydshln}
\usepackage{tabularx}
\usepackage{verbatim}
\usepackage{CJK}
\usepackage{flushend}
\usepackage{multicol}
\usepackage{multirow,makecell}
\usepackage{color}
\usepackage{colortbl,array}
\usepackage{silence}
\usepackage{arydshln} 
\usepackage{xspace}
\usepackage{soul}
\usepackage{pifont}

\usepackage{arydshln}

\usepackage{amsmath,amsfonts,bm}

\def\Figref#1{{Fig.}~\ref{#1}}

\def\Secref#1{{\S}\ref{#1}}

\def\eqref#1{equation~\ref{#1}}
\def\Eqref#1{{Eq.}~(\ref{#1})}

\def\Tabref#1{{Tab.}~\ref{#1}}

\def\1{\bm{1}}

\def\vs{{\bm{s}}}

\DeclareMathAlphabet{\mathsfit}{\encodingdefault}{\sfdefault}{m}{sl}
\SetMathAlphabet{\mathsfit}{bold}{\encodingdefault}{\sfdefault}{bx}{n}

\newcommand{\E}{\mathbb{E}}

\newcommand{\R}{\mathbb{R}}

\DeclareMathOperator*{\argmax}{arg\,max}

\usepackage{tikz}

\makeatletter
\DeclareRobustCommand\onedot{\futurelet\@let@token\@onedot}
\def\@onedot{\ifx\@let@token.\else.\null\fi\xspace}

\def\eg{\textit{e.g}\onedot} 
\def\ie{\textit{i.e}\onedot} 
 
 \def\vs{\textit{vs}\onedot}

\def\wrt{\textit{w.r.t}\onedot} 

\makeatother

\newcommand{\hidethis}[1]{}

\definecolor{bblue}{HTML}{4F81BD}
\definecolor{oorange}{HTML}{F4C842}
\definecolor{rred}{HTML}{C0504D}
\definecolor{ggreen}{HTML}{9BBB59}
\definecolor{ppurple}{HTML}{9F4C7C}
\definecolor{darkgreen}{HTML}{228B22}
\definecolor{cred}{HTML}{D81B60}
\definecolor{cblue}{HTML}{1E88E5}
\definecolor{cyellow}{HTML}{FFC107}
\definecolor{nred}{HTML}{e41a1c}
\definecolor{nblue}{HTML}{377eb8}
\definecolor{ngreen}{HTML}{4daf4a}
\definecolor{lblue}{HTML}{6C8EBF}

\newlength\savewidth

\newcolumntype{x}[1]{>{\centering\arraybackslash}p{#1pt}}
\newcolumntype{y}[1]{>{\raggedright\arraybackslash}p{#1pt}}
\newcolumntype{z}[1]{>{\raggedleft\arraybackslash}p{#1pt}}

\newcommand{\app}{\raise.17ex\hbox{$\scriptstyle\sim$}}

\definecolor{deemph}{gray}{0.6}

\definecolor{baselinecolor}{gray}{.9}

\newcommand{\smallcitep}[1]{\footnotesize\citep{#1}}

\usepackage{color, colortbl}
\definecolor{emerald}{rgb}{0.31, 0.78, 0.47}
\definecolor{Gray}{gray}{0.9}
\definecolor{Highlight}{rgb}{0.89,0.89,0.94}
\usepackage[first=0,last=9]{lcg}
\newcommand{\chl}{\cellcolor{Highlight}}

\renewcommand{\paragraph}[1]{\noindent\textbf{#1}}
\renewcommand{\bm}[1]{\mathbf{#1}}

\newcommand{\method}{\textsc{DPLM}\xspace}

\newcommand{\pLM}{\textit{p}\textsc{LM}\xspace}
\newcommand{\pLMs}{\textit{p}\textsc{LM}s\xspace}
\newcommand{\MLM}{Masked-LM\xspace}
\newcommand{\ARLM}{\textsc{Ar-LM}\xspace}

\newcommand{\xt}[1]{\bm{x}^{(#1)}}
\newcommand{\metric}[1]{\texttt{#1}}

\begin{document}

\twocolumn[

\icmltitle{Diffusion Language Models Are Versatile Protein Learners}


\icmlsetsymbol{equal}{*}

\begin{icmlauthorlist}
\icmlauthor{Xinyou Wang}{equal,nju,byted}
\icmlauthor{Zaixiang Zheng}{equal,byted}
\icmlauthor{Fei Ye}{byted}
\icmlauthor{Dongyu Xue}{byted}
\icmlauthor{Shujian Huang}{nju}
\icmlauthor{Quanquan Gu}{byted}
\end{icmlauthorlist}

\icmlaffiliation{byted}{ByteDance Research}
\icmlaffiliation{nju}{Dept. of Computer Science, Nanjing University (this work was done during Xinyou's internship at ByteDance Research)}

\icmlcorrespondingauthor{Quanquan Gu}{quanquan.gu@bytedance.com}

\icmlkeywords{Machine Learning, ICML}

\vskip 0.3in
]



\printAffiliationsAndNotice{\icmlEqualContribution} 


\begin{abstract}

This paper introduces {\ul{\textbf{d}}iffusion \ul{\textbf{p}}rotein \ul{\textbf{l}}anguage \ul{\textbf{m}}odel} (\method), a versatile protein language model that demonstrates strong generative and predictive capabilities for protein sequences.
We first pre-train scalable {\method}s from evolutionary-scale protein sequences within a generative self-supervised discrete diffusion probabilistic framework, which generalizes language modeling for proteins in a principled way. 
After pre-training, \method exhibits the ability to generate structurally plausible, novel and diverse protein sequences for unconditional generation.
We further demonstrate the proposed diffusion generative pre-training make \method possess a better understanding of proteins, making it a superior representation learner, which can be fine-tuned for various predictive tasks, comparing favorably to ESM2~\citep{lin2022esmfold}.
Moreover, \method can be tailored for various needs, which showcases its prowess of conditional generation in several ways: (1) conditioning on partial peptide sequences, \eg, generating scaffolds for functional motifs with high success rate; (2) incorporating other modalities as conditioners, \eg, structure-conditioned generation for inverse folding; and 
(3) steering sequence generation towards desired properties, \eg, satisfying specified secondary structures, through a plug-and-play classifier guidance.
Code is released at \url{https://github.com/bytedance/dplm}.

\end{abstract}

\vspace{-7mm}
\section{Introduction}
\label{sec:intro}

Proteins, which are 3D-folded linear sequences of amino acids, play a pivotal role in regulating various biological functions, including transcription, translation, signaling, and the control of the cell cycle.
Recently, the promise of learning to understand and design proteins via data-driven generative deep learning has initiated a significant paradigm shift apart from the long-established physics-based methods.

The analogies between protein sequences and human languages have long been recognized~\cite{yang2019machine,ferruz2022controllable}.
Drawing inspiration from the remarkable progress in NLP achieved by language models~\citep[LMs;][]{devlin2019bert,radford2018gpt,openai2023gpt4} thanks to the \textit{scalability} of Transformers~\citep{vaswani2017attention} and the existence of large-scale text data, recent explorations in protein has also demonstrated the impressive capabilities of protein language models~\citep{rives2019esm,lin2022esmfold,hu2022exploring}, learned from the universe of evolutionary-scale protein sequences.
As a result, protein LMs have become one of the most important cornerstones in AI for protein research, serving a pivotal role not only in predictive tasks (\eg, probing functional properties, and predicting protein structures from single sequences without explicit evolutionary homologs) but also in generative tasks (\eg, redesigning sequences given protein backbone structures, or synthesizing completely new protein sequences). 

\begin{figure*}[t]
    \vspace{-1.5mm}
    \centering
    \includegraphics[width=0.96\linewidth]{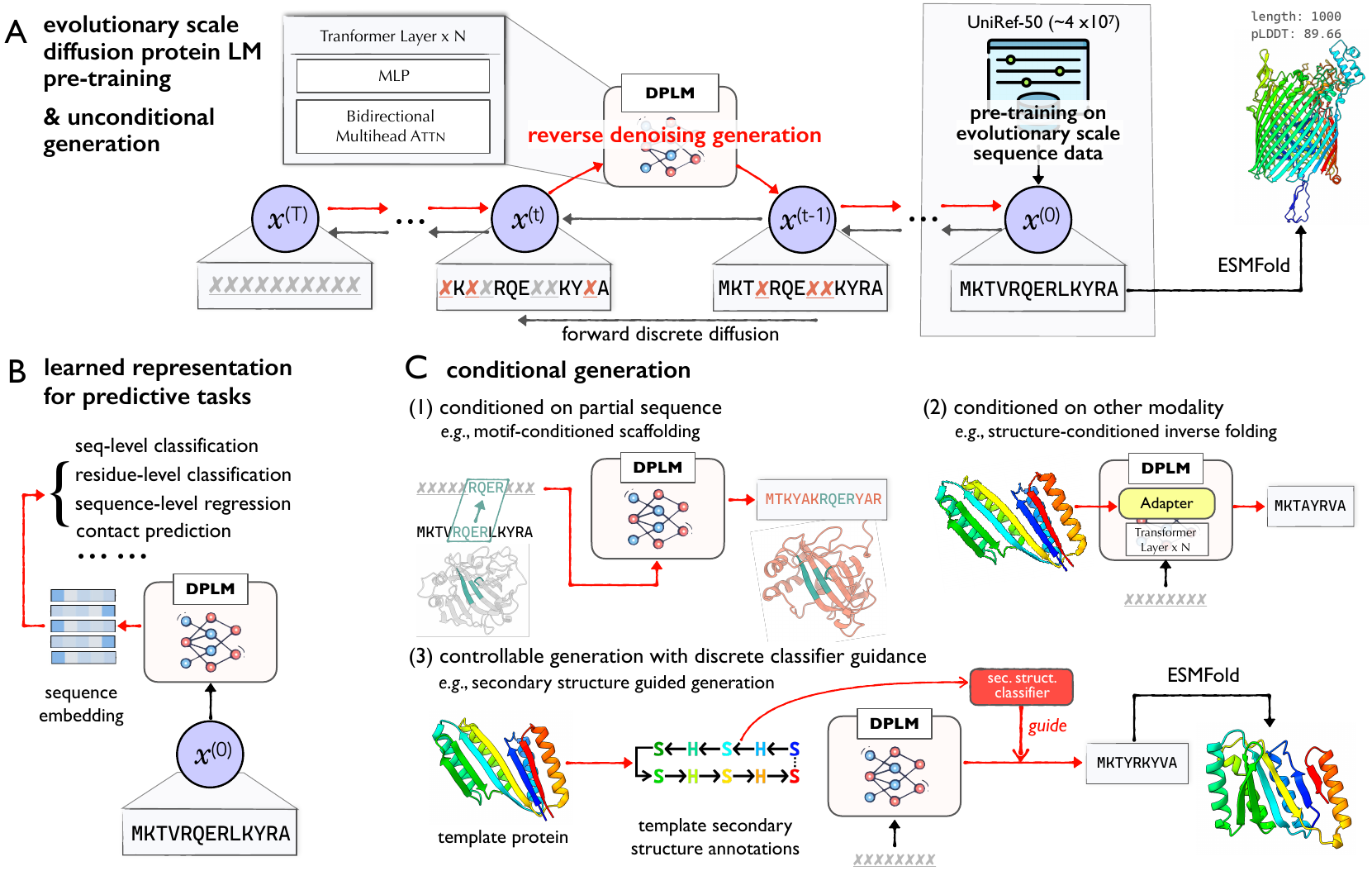}
    \vspace{-4mm}
    \caption{{\sl Overall illustration of \method.}
    \textbf{(A)}: modeling, pre-training and unconditional generation; 
    \textbf{(B)}: protein sequence representation for predictive tasks; 
    \textbf{(C)}: conditional generation, including \textbf{(1)} sequence conditioning (\eg, motif-scaffolding), \textbf{(2)} cross-modal conditioning (\eg, inverse folding), and \textbf{(3)} plug-and-play controllable generation with discrete classifier guidance (\eg, secondary structure).}
    \label{fig:main}
    \vspace{-2mm}
\end{figure*}

While current protein LMs have made significant strides, they have not yet reached their fullest potential.
One of the fundamental problems is rooted in the widely-used pretraining objectives, \ie, masked prediction \vs autoregression:
\begin{compactitem}
    \item[(i)] For masked prediction, masked language models \citep[{\MLM}s, \eg, ESM family;][]{rives2019esm,lin2022esmfold} excel in sequence understanding for protein predictive tasks, thanks to their \textit{bi-directional receptive field}. 
    However, {\MLM}s are unable to perform protein sequence generation, due to the lack of a well-defined formulation for generative modeling. 
    We further postulate that this could even cap their predictive power, since a powerful generative model that can \textit{create} new samples by learning the underlying data distribution, is expected to simultaneously acquire a deep \textit{understanding} of the data.
    As a famous quote, ``\textit{what you cannot create, you do not understand}.''
    
    \item[(ii)] For autoregression, autoregressive language models~\citep[{\ARLM}s, \eg, ProGen;][]{nijkamp2022progen2}, albeit good at generation, often fall short in understanding sequence data~\citep{radford2018gpt} including proteins~\citep{elnaggar2021prottrans}. 
    More importantly, proteins are structural macromolecules rather than simple linear strings. 
    Consequently, while effective as an inductive bias for text, {\ARLM}s are constrained by their uni-directional receptive field, only accessing one-sided sequence context. 
    This limitation stems from capturing the complex global interactions of amino acids, thereby hindering both generative and predictive capabilities of protein LMs.
\end{compactitem}
This highlights the demand for a general-purpose and versatile protein LM that combines predictive and generative capabilities.
Provided the aforementioned analysis, we reason that, \textit{the key ingredients} for such a versatile protein LM lie in (1) \textit{strong \& scalable generative modeling framework} to best digest the universe of massive protein sequences; and (2) \textit{bi-directional receptive field} for better modeling residue-wise global interactions. 

\vspace{1pt}
On the other hand, diffusion models~\citep{ho2020ddpm,song2020sde} have shown great success in generating \textit{continuous} data, especially in rendering photorealistic images~\citep[][\textit{inter alia}]{rombach2021highresolution}.
They have further manifested incredible achievement in modeling protein structures~\citep{yim2023framediff,watson2023RFdiffusion,ingraham2023chroma}.
This can be attributed to their favorable properties of non-autoregressive denoising generation with iterative refinement and global receptive field.
Besides, denoising autoencoding has a long history for representation learning~\citep[][\textit{inter alia}]{vincent2010stacked}, while recent studies have verified that diffusion-based generative models can be effective self-supervised learners~\citep{chen2024deconstructing}.
These make diffusion models an appealing generative foundation for protein language modeling. 
However, directly applying conventional Gaussian diffusion to protein sequences necessitates additional continuous relaxations~\citep{lisanza2023joint}, which does not fit the \textit{discrete} nature of protein sequence and has not yet proven successful in practice.


In this paper, 
we present {\ul{\textbf{d}}iffusion \ul{\textbf{p}}rotein \ul{\textbf{l}}anguage \ul{\textbf{m}}odel} (\method), a novel approach aimed at achieving a unified and versatile protein LM through diffusion generative pre-training on evolutionary-scale protein sequences. 
\method is grounded in a discrete diffusion probabilistic framework, serving as a principled generative generalization of language modeling.
During pre-training, \method is tasked with denoising the input protein sequence at different noise levels, ranging from completely noisy to clean ones, enforcing \method to best the model complex intrinsic dependencies of amino acid sequences.
After pre-training, \method can be used for protein sequence generation and providing effective representations for downstream predictive tasks.
We highlight our contributions as follows:
\begin{compactitem}
    \item We propose \method, a versatile protein LM under discrete diffusion framework, with model size up to 3B, pre-trained on evolutionary-scale protein sequences. 
    We further develop multiple conditioning strategies covering various use needs, especially discrete classifier guidance for controllable generation.
    As a result, \method combines the best of both worlds, \ie, the scalable expressiveness of language models and the strong generative power of diffusion models, serving as a versatile biological foundation model
    (\Figref{fig:main}, \Secref{sec:method}).
    
    \item We show that \method is capable of generating highly structurally plausible (\ie, averaged \metric{pLDDT}~$>80$), novel and diverse for unconditional protein sequence generation, suggesting that \method well captures the universe of protein sequence data (\Figref{fig:main}A, \Secref{sec:uncond}). 

    \item We demonstrate that \method \textit{understands} protein better, serving as a superior representation learner, which can be fine-tuned for various downstream tasks, comparing favorably with widely-used protein sequence encoder models, \eg, ESM-2~\citep{lin2022esmfold} (\Figref{fig:main}B, \Secref{sec:understanding}). 
    
    \item \method can be further exploited for conditional generation for a variety of needs:
    \method can (1) condition on pre-specified partial sequence, \eg, scaffolding for functional motifs with high success rate;
    (2) incorporate other modalities as conditions, \eg, structure-conditioned generation for inverse folding; 
    (3) generate protein sequences towards desired properties with plug-and-play classifier-guidance, \eg, steering \method to synthesize proteins that satisfy arbitrary user-defined secondary structure annotations (\Figref{fig:main}C, \Secref{sec:exp_cond}).

\end{compactitem}

\section{Preliminaries}
\label{sec: preliminary}

\subsection{Language Modeling for Protein} 
Language modeling aims to estimate the underlying distribution $\bm{x} \sim q(\bm{x})$ of the sequence data of our interest, \eg, text or protein sequence, by learning a probabilistic model $p_\theta(\bm{x})$.
Here the \textit{language model} (LM) $\theta$ is parameterized by a neural network, in particular Transformers~\citep{vaswani2017attention}, which have become the \emph{de facto} choice dominating different domains with scalable and performing expressiveness.
In this work, we are interested in language modeling for protein sequences, for which $\bm{x}=(x_1, x_2, \dots, x_L) \in \{0,1\}^{L \times |\mathcal{V}|}$ is a sequence composing $L$ elements, $\mathcal{V}$ is the vocabulary within a discrete data support of 20 amino acids $\mathcal{V} = \{1,..., 20\}$.
One thing we most care about is the generative and representational capabilities of protein LMs.
Here we review the typical probabilistic paradigms for language modeling, \ie, \textit{masked prediction }and \textit{autoregression}, and their pros and cons as the foundation for protein LMs, as follows.

\paragraph{Masked Prediction.} 
Masked language models ({\MLM}s or MLMs), \eg, BERT~\cite{devlin2019bert} and its variants for protein sequence~\citep[ESM family,][]{rives2019esm,lin2022esmfold}, employ a bidirectional transformer to take into account both the left and right context to predict the masked (amino acid) symbols in a \textit{mask-predict} autoencoding manner,
\begin{align}
   \E_{q(\bm{x})} \log p_\theta(\bm{x}) = \E_{q(\bm{x})} \textstyle{\sum_{1 \leq i \leq L}} b_i \cdot \log p_\theta(x_i | {\Bar{\bm{x}}_{\text{m}}}), 
   \label{eq:mlm}
\end{align}
where $b_i = \bm{1}_{\Bar{\bm{x}}_i = \texttt{[X]}}$ derived from a fixed chance (\eg, widely-adopted $15\%$) of masking $\bm{x}$ with a special mask symbol $\texttt{[X]}$, resulting in the masked observation $\Bar{\bm{x}}_{\text{m}}$.
A per-token conditional independence assumption is made as well.
{\MLM}s significantly excel the performance of a wide range of sequence understanding tasks for both natural language and protein.
However, its bidirectionality nature makes it difficult to apply to sequence generation.

\paragraph{Autoregression.} 
{\ARLM}s are prevailing in the realm sequence generation~\citep{openai2023gpt4,nijkamp2022progen2}, which adopts a sequential factorization over the sequence using the probability chain rule. 
In this case, the log-likelihood of such models is maximized over the dataset given by:
\begin{align}
   \E_{q(\bm{x})} \log p_\theta(\bm{x}) = \E_{q(\bm{x})} \textstyle{\sum_{1\leq i \leq L}} \log p_\theta(x_i | \bm{x}_{<i}),
   \label{eq:arlm}
\end{align}
where causal masking is used to ensure sequential dependency structure.
To sample from AR-LMs, it requires ancestral sampling for $L$ iterative steps from $x_1 \sim p_\theta(x_1), x_2 \sim p_\theta(x_2 | x_1)$ towards $x_L \sim p(x_L | x_1, ..., x_{L-1})$ in a strict left-to-right unidirectional manner.

\subsection{Diffusion Probabilistic Models}
Diffusion models~\citep{sohl2015diffusion,ho2020ddpm,song2020sde} are a class of generative models characterized by a pair of Markov processes, \ie, a forward diffusion process and a backward denoising process. 
The \textit{forward} process~${q(\xt{1:T}|\xt{0})=\prod_{t=1}^T q(\xt{t}|\xt{t-1})}$ gradually perturb the data $\xt{0}\sim q(\xt{0})$ into a stationary distribution $\xt{T} \sim q_{\text{noise}}$ with $T$ increasingly noisy steps $\xt{1:T}=\bm{x}_{1}, \dots, \xt{t-1},\xt{t}, \dots, \xt{T}$. 
The learned \textit{backward} process ${p_{\bm{\theta}}(\xt{0:T}) = p(\xt{t})\prod_{t=1}^{T}p_{\bm{\theta}}(\xt{t-1}|\xt{t})}$, reversely, gradually denoises the samples towards the data distribution. 
To fit the model $p_{\bm{\theta}}(\xt{0})$ to the data distribution $q(\xt{0})$, the denoiser model is typically optimized by the variational bound of the log-likelihood~\citep{ho2020ddpm}:
\begin{align}
      & \mathbb{E}_{q(\xt{0})}\big[\log p_\theta(\xt{0})\big]  \geq \mathbb{E}_{q(\xt{0:T})} \bigg[\log \frac{p_{\theta}(\xt{0:T})}{q(\xt{1:T}|\xt{0})}\bigg] \nonumber\\[-5pt]
      & = \mathbb{E}_{q(\xt{0})}\Big[\log p_{\theta} (\xt{0} | \xt{1}) +\text{const.}  \nonumber \\[-5pt]
      & \qquad + \textstyle{\sum_{t=2}^{T}} \underbrace{-\text{KL}\big[q(\xt{t-1}|\xt{t}, \xt{0})\|p_{{\theta}}(\xt{t-1}|\xt{t})\big]\Big]}_{\mathcal{J}_t}. \nonumber 
\label{eqn: variational bound}
\end{align}
Afterwards, it generates by first sampling from $q_{\text{noise}}(\xt{T})$, followed by iterative denoising with $p_{{\theta}}(\xt{t-1}|\xt{t})$.

\section{\method: A \textit{Versatile} Protein LM}
\label{sec:method}

\paragraph{Motivation.}
Continuous diffusion with Gaussian perturbation kernel 
has demonstrated impressive performance in generating continuous data in Euclidean space~\citep{rombach2021highresolution, ho2022imagenvideo}, and the more general Riemannian manifolds~\citep{de2022riemannian}. 
Recently, continuous diffusion has shown to rival in modeling protein structures~\citep[][\textit{inter alia}]{watson2023RFdiffusion,ingraham2023chroma}, wherein its \textit{bidirectional receptive field} is ideally suited for modeling residue-wise global interactions. 
This motivates us to blend diffusion models, which are well-suited for protein as discussed above, and language models, which are well known as \textit{scalable and expressive sequence learners}.
This leads to our pursuit of a diffusion protein LM, taking the best of both worlds.

A direct use of continuous diffusion, however, is not necessarily the best choice for modeling discrete sequence data~\citep{li2022diffusionlm,cdcd,lisanza2023joint}, due to the \textit{pitfall of discreteness} that makes Gaussian diffusion hardly model the discrete nature of sequence data in embedding space~\citep{ye2023dinoiser}.
To this end, discrete diffusion~\citep{hoogeboom2021argmax,austin2021structured} that directly operates over the discrete state space, becomes a more well-suited probabilistic model for protein sequences.



\subsection{Protein Language Modeling \textit{w/} Discrete Diffusion}
\label{sec:modeling}

\paragraph{Modeling.}
Let $\texttt{Cat}(\bm{x};\bm{p})$ be a categorical distribution on protein sequence $\bm{x}$ parameterized by a vector $\bm{p}$ on $(|\mathcal{V}|-1)$-dimensional probability simplex. The forward process of discrete diffusion defines a Markov process governed by the transition kernel: 
\begin{equation}
q(\xt{t}|\xt{t-1})=\texttt{Cat}\big(\xt{t}; \beta_t\xt{t-1} + (1-\beta_t)\bm{q}_{\text{noise}}\big), \nonumber
\end{equation}
where $\bm{q}_\text{noise}$ is the probability vector of stationary distribution $q_\text{noise}(\xt{t})$, \ie, $q(\xt{t})=\texttt{Cat}(\xt{t}; \bm{p}=\bm{q}_{\text{noise}})$, and $0\ll\beta_t<1$ is the noise schedule controlling the degree of corruption at timestep $t$.
In this case, the distribution of corrupted sample $\xt{t}$ given its original data $\xt{0}$ has a closed-form expression:
\begin{equation}
q(\xt{t}|\xt{0})= \texttt{Cat}\big(\xt{t}; \alpha_t\xt{0}+ (1-\alpha_t)\bm{q}_{\text{noise}}\big),
\label{eqn: dd 0_to_t}
\end{equation}
where $\alpha_t = \prod_{i=1}^t\beta_i$ such that $\lim_{t \to T} \alpha_t \to 0$, which preserves no information from the data and converges to the stationary distribution $\bm{q}_{\text{noise}}$ at timestep $T$.
This shows that the diffusion process is intuitively a convex combination between data and the stationary noise prior distribution.
Different stationary distributions $\bm{q}_\text{noise}$ lead to different formulations of discrete diffusion models. 
Here we primarily consider the \textit{absorbing} diffusion with $q(\xt{t}) = \{ 1~\text{if}~ \xt{t} = \texttt{[X]};~0 ~\text{if}~ \xt{t} \not= \texttt{[X]} \}$, where \texttt{[X]} is an absorbing state, akin to {\MLM}s. 
The formulation of \Eqref{eqn: dd 0_to_t} results in $\xt{t}$ either being masked or the same as $\xt{0}$, with a masking ratio $(1-\alpha_t)$. 

\paragraph{Learning.} 
As stated in \citet{austin2021structured}, discrete diffusion inherently connects to AR-LM and Masked-LM, whilist \citet{zheng2023reparameterized} further simplifies the learning objective of discrete diffusion, with their proposed reparameterized backward transition, from KL divergences between two categoricals into reweighted cross-entropies:
\begin{align}
\mathcal{J}_t & = \mathbb{E}_{q(\xt{0})}-\text{KL}\big[q(\xt{t-1}|\xt{t}, \xt{0})\|p_{{\theta}}(\xt{t-1}|\xt{t})\big] \nonumber \\[-1pt]
& = \mathbb{E}_{q(\xt{0})} \Big[\lambda^{(t)}  \textstyle{\sum_{1 \leq i \leq L}} b_i(t) \cdot \log p_{\theta}(\xt{0}_i|\xt{t})\Big], 
\label{eq:reparam_obj}
\end{align}
where $\lambda^{(t)}$ is a weighting coefficient induced from the specific noising schedule (see Appendix~\ref{app: rdm} for proof).
\Eqref{eq:reparam_obj} reveals that 
{\MLM}s~(\ie, $\xt{t}\triangleq\Bar{\bm{x}}_{\text{m}}$ in \Eqref{eq:mlm}) and {\ARLM}s~(\ie, $\xt{t}\triangleq\bm{x}_{<t}$ and $b_i \triangleq 1 $ in \Eqref{eq:arlm}) can be considered as special cases in this generalized form of discrete diffusion LMs, contingent on their respective specifications of the noise-induced configurations.
As a result, the process of learning according to \Eqref{eq:reparam_obj} inherently encapsulates both {\MLM}s and {\ARLM}s within the ambit of the proposed \method.


\paragraph{Evolutionary-scale Pre-training.}
The pre-training procedure for \method utilizes the UniRef50 database~\citep{suzek2015uniref}, which comprises around 45 million protein sequences, totaling about 14 billion amino acid tokens. 
In the case of exceedingly lengthy protein sequences, we emulate ESM2~\citep{lin2022esmfold} by truncating these proteins to a random sequence of 1024 tokens. 
Besides, we adhere to the setting for model architecture and scales as ESM2, which correspond to \method with sizes of 150M, 650M and 3B.
We train all models for 100K updates, with batch size of 320K for 150M model and 1M for 650M/3B models.

\paragraph{Generation.}
Given a trained \method, it can synthesize new amino acid sequences by the reverse iterative denoising process of discrete diffusion~\citep{hoogeboom2021argmax,austin2021structured}.
Formally, discrete diffusion samples from the following distribution,
\begin{align}
    p_\theta(\xt{t-1} | \xt{t}) = \textstyle{\sum_{\hat{\bm{x}}_0}}  q(\xt{t-1} |\xt{t}, \hat{\bm{x}}_0) p_\theta(\hat{\bm{x}}_0 | \xt{t}). \nonumber
\end{align}
In particular, at time $t$, we first generate $\hat{\bm{x}}_0$ from $p_\theta(\cdot| \xt{t})$, then a less noisy $\xt{t-1}$ is sampled by $q(\cdot |\xt{t},\xt{0} = \hat{\bm{x}}_0)$ given $\xt{t}$ and $\hat{\bm{x}}_0$. This process is repeated from $T$ to $1$.
The generative denoising process of \method can be viewed as an iterative \textit{mask-predict} approach.
Specifically, the starting sequence is initialized as $100\%$-noisy state (\ie, all $\texttt{[X]}$'s).
At each iteration, a subset of masked tokens is updated based on the model's prediction $\hat{\bm{x}}_0$, while the remaining tokens are re-masked, according to ranked $\log p_\theta(\hat{\bm{x}}_0 | \xt{t})$~\citep{ghazvininejad2019mask,zheng2023reparameterized}.

\paragraph{Representation.}
\method is tasked with denoising the input protein sequence at all noise levels, including the original noise-free data (\eg, noise level at $0\%$).
As a result, \method can simultaneously serve as a protein sequence representation learner over massive protein sequence data, providing useful sequence embedding for various protein predictive downstream tasks, \eg, sequence/residue-level classification/regression.
The sequence embedding can be attained by simply letting \method take as input the given amino acid sequence $\bm{x}$: $\bm{h}(\bm{x}) \leftarrow \method_\theta(\bm{x}, t=0) \in \R^{L \times d}$, where $d$ is the dimension of embedding.


\subsection{Conditioning}
Being able to efficiently sample realistic proteins is necessary but not sufficient for downstream applications such as therapeutic development, since unconditional samples are unlikely to possess desired functional properties.
Here we elaborate on how to make \method practically useful by conditioning for various needs, which covers most common scenarios, \ie, sequence conditioning, cross-modal conditioning, and plug-and-play preference-guided conditioning.

\paragraph{Case I: Conditioning on partial sequence (\Figref{fig:main}C-1).}
Protein generation containing pre-specified polypeptides corresponds to various use cases such as generating scaffolds for given functional motifs, infilling antibody CDR loops, or imposing expert knowledge a-priori.
This implies our desire for \method to sample from this conditional distribution $\bm{x} \sim p_\theta(\bm{x} | \bar{\bm{x}}) = \prod_{i=1}^L b_i \cdot p_\theta(x_i | \Bar{\bm{x}})$, which has already been learned through~\Eqref{eq:reparam_obj}.
The observed partial sequence $\Bar{\bm{x}} = \{ \bar{x}_i \in \mathcal{V}~\text{if}~ b_i = 0; \texttt{[X]}~\text{if}~ b_i = 1 | i \in [1,L]\}$. 
Namely, $b_i \in \{0,1\}$ indicates whether the predicted sequence must preserve the observation for the $i$-th residue such that $x_i = \Bar{x}_i$. 

\paragraph{Case II: Adapting \method to conditioned on other modalities (\Figref{fig:main}C-2).}
Generating protein sequence subject to cross-modal constraints $\bm{c}$, \ie, $\bm{x} \sim p_{\theta}(\bm{x} | \bm{c})$, has profound value in practice, such as inverse protein folding where sequences are generated for given backbone structure~\citep{dauparas2022proteinmpnn,zheng2023structure}, or conditioning on small molecule ligands for binder design~\citep{dauparas2023atomic}.
Given that \method primarily operates over amino acid tokens,
in these cases, we can equip \method with cross-modal conditioning by adapter-tuning with a pre-trained modality expert encoder $\mathcal{E}_\phi(\bm{c})$ and a newly-added cross-attention-based adapter following \citet{zheng2023structure}.
During training, we freeze the parameters of the modality encoder and \method, and only update the parameters of the adapter via supervised fine-tuning on the given paired data $(\bm{x}, \bm{c})$.
We then obtain a conditional \method for $p_{\theta}(\bm{x} | \mathcal{E}_\phi(\bm{c}))$ making the full potentials of both \method and the modality expert $\mathcal{E}_\phi(\bm{c})$.
In \Secref{sec:classifier-free}, we also develop classifier-free guidance for such adapter-tuned \method as an immediately available booster for cross-modal conditional generation without intricate condition dropout during training.

\paragraph{Case III: Plug-and-play controllable generation with discrete classifier guidance (\Figref{fig:main}C-3).}
Directly building a conditional model is prohibitive in most cases due to data scarcity.
Thus, incorporating classifier guidance into continuous diffusion models~\citep{dhariwal2021diffusion} proves particularly useful. This integration with pre-trained classifiers enables steering generation towards desired preferences.
However, continuous classifier guidance requires valid definition of $\nabla_{\bm{x}} \log p_\theta(\bm{x})$, or ``score''~\citep{song2019score}, which does not exist for discrete diffusion.
Inspired by continuous diffusion classifier guidance and DiGress on guided graph diffusion~\citep{vignac2022digress}, here we introduce classifier-guided conditional generation for discrete diffusion LMs. 
Concretely, we want to sample from the conditional distribution of $q(\xt{t-1}| \xt{t}, \bm{y}) \propto q(\xt{t-1}| \xt{t}) q(\bm{y} |\xt{t-1} )$, which is approximated by $ p_\theta(\xt{t-1}| \xt{t}) p_\phi(\bm{y} |\xt{t-1})$ where $p_\phi(\bm{y} |\xt{t-1})$ is a discriminative guidance model (classifier or regressor \wrt user's desired properties). 
However, $p_\phi(\bm{y} |\xt{t-1})$ cannot be factorized as a product over all positions, prohibiting evaluation of all possible values of $\xt{t-1}$. 
To this end, we resort to an approximation with first-order Taylor expansion around $\xt{t}$~\citep{dhariwal2021diffusion}, where we treat $\bm{x}$ as a continuous one-hot variable on probability simplex to make $\nabla_{\bm{x}}$ a valid operator, thereby,
\begin{align}
    & \log q(\bm{y} |\xt{t-1}) \nonumber \\[-1pt]
    &\approx~  \log q(\bm{y} |\xt{t}) + \langle \nabla_{\bm{x}} \log q(\bm{y} | \xt{t}), \xt{t-1} - \xt{t} \rangle \nonumber \\
    &\approx~  \textstyle{\sum_{1\leq i \leq L}} \langle \nabla_{\bm{x}_{i}} \log q(\bm{y} | \xt{t}), \bm{x}_{i}^{(t-1)} \rangle + C(\xt{t})\nonumber,
\end{align}
where $C(\xt{t})$ is a constant that does not depend on $\xt{t-1}$. 
We use $p_\phi(\bm{y} |\xt{t})$ to estimate $ q(\bm{y} |\xt{t})$ and plug it into the above expression. 
We can now sample from the resulting conditional distribution instead at each timestep $t$,
\begin{align}
    \xt{t-1} & \sim p_\theta(\xt{t-1}| \xt{t}) p_\phi(\bm{y} | \xt{t-1})^{\eta} \label{eq:classifier-guidance}   \\ 
    & \propto p_\theta(\xt{t-1}| \xt{t}) e^{\big( \eta \cdot \sum_{i} \langle \nabla_{\bm{x}_{i}} \log p_\phi(\bm{y} | \xt{t}), \bm{x}_{(i)}^{t-1} \rangle\big)}, \nonumber
\end{align}
where a tunable $\eta$ controls the strength of guidance.

\begin{figure*}[t!]
    \centering
    \vspace{-3mm}
    \includegraphics[width=0.97\linewidth]{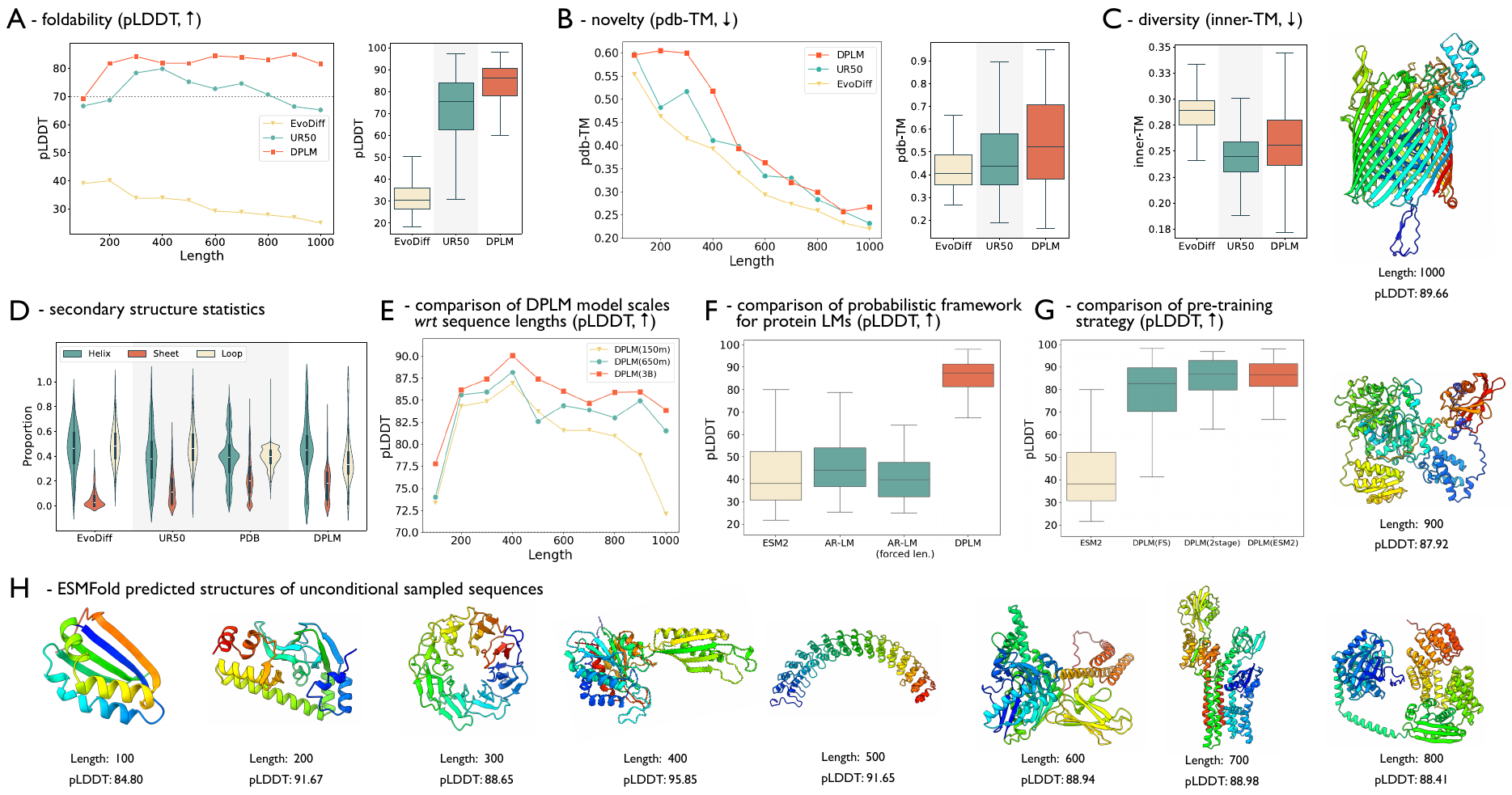}
    \vspace{-4mm}
    \caption{\emph{Evaluation of unconditional generation.} Here we use ESMFold as the folding model to predict structures and calculate \metric{pLDDT} for all the sampled sequences. 
    We measure the (structural) novelty of the generated sequences against all known structures in PDB by \metric{TM-score} (\ie, \metric{pdb-TM}, and measure the (structural) diversity within the sampled candidates for each model (\ie, \metric{inner-TM}).
    }
    \label{fig:uncond_sample}
    \vspace{-3mm}
\end{figure*}

\subsection{Comparisons with The Most Related Work}

Comprehensive representations for protein sequence understanding are achieved by pre-training on protein sequence data via masked language modeling~\citep{devlin2019bert}, akin to language understanding.
Among those, the family of ESM-1b/ESM2~\citep{rives2019esm,lin2022esmfold} serves as the pioneer \& cornerstone sequence embedding models for extensive protein predictive tasks. Therefore, \method follows the best practice of ESM2 in network architecture and pre-training strategies. 
\method takes a significant leap from ESM2 with immediate strong generative capabilities, without expensive needs for Monte Carlo methods~\citep{verkuil2022language} or Gibbs sampler~\citep{johnson2021generating}, which treat \MLM as Markov random fields~\citep{wang2019bert}. 
Besides, as verified from predictive experiments (\Secref{sec:understanding}), the generative ability of \method further enables its enhanced representation learning, echoing Richard Feynman's famous quote ``\textit{What I cannot create, I do not understand}''.

Regarding protein sequence generation, EvoDiff~\citep{alamdari2023protein} is the most relevant approach, which uses order-agnostic autoregressive diffusion models~\citep[OADM,][]{hoogeboom2021autoregressive} for unconditional generation, with conditional applications on intrinsic disordered sequence infilling and motif-scaffolding, whereas attaining better performance necessitates multiple sequence alignments (MSAs) based on a MSA-Transformer~\citep{rao2021msa_trans} parameterization. 
\method differs from EvoDiff in several aspects:
(1) \method manifests superior representation learning, which, to the best of our knowledge, is the first time for protein diffusion models, even in general language learning regime, showing \method's appealing versatility, as shown in \Tabref{tab:results_understanding}; 
(2) \method is based on a more principled discrete diffusion framework beyond the special (order-agnostic) autoregressive diffusion, which is not compatible with refining intermediate predictions and requires expensive $O(L)$ decoding overhead;
(3) we investigate the ability of \method to accommodate extensive conditioning, especially conditioning on other modality and programmable generation steered by discrete classifier guidance, pushing steps forward beyond simple sequence conditioning investigated in EvoDiff paper.

Please refer to Appendix \Secref{sec:related} for a more detailed discussion of the related work.



\section{Experiments}
\label{sec:experiment}

We evaluate \method on extensive generative and understanding tasks, spanning unconditional generation (\Secref{sec:uncond}), a variety of protein predictive downstream tasks (\Secref{sec:understanding}), and conditional tasks, including motif-scaffolding (\Secref{sec:motif}), inverse-folding task (\Secref{sec:IF}), and secondary structure guided controllable generation (\Secref{sec:controllable}). 
We find that, in general, \method with larger model scales can attain better results than smaller ones, demonstrating the scaling law can also hold for protein language modeling.
Please refer to the Appendix for more detailed experimental settings.

\begin{table*}[t!]
\centering
\footnotesize
\vspace{-4mm}
\setlength{\tabcolsep}{4.5pt}
\caption{{\sl Performance on various protein predictive downstream tasks.}
$\dagger$: benchmarked results are quoted from \citet{su2023saprot}.
}
\label{tab:results_understanding}
\vspace{1.5pt}
\resizebox{\linewidth}{!}{%
\begin{tabular}{lcccccccccc}
\toprule
\multirow{3}{*}{{Models}} & \multirow{2}{*}{Thermostability} & \multirow{2}{*}{HumanPPI} & \multirow{2}{*}{Metal Ion Binding} & \multirow{2}{*}{EC} & \multicolumn{3}{c}{GO}                           & \multicolumn{2}{c}{DeepLoc}   & \multirow{2}{*}{SSP}  \\
\cmidrule[0.5pt](lr){6-8}  \cmidrule[0.5pt](lr){9-10} &  &  &  &  & MF & BP & CC & Subcellular   & Binary  & CASP12 \\
\cmidrule[0.5pt](lr){2-11} 
    & \metric{Spearman's} $\rho$      & \metric{Acc} ($\%$)       & \metric{Acc} ($\%$)    & \metric{Fmax}      & \metric{Fmax}   & \metric{Fmax}    & \metric{Fmax}    & \metric{Acc} ($\%$)   & \metric{Acc} ($\%$)     & \metric{Acc} ($\%$) \\
\midrule
$^\dagger$SaProt (*structure provided)    & 0.724    & 86.41      & 75.75    & 0.884      & 0.678    & 0.356 & 0.414    & 85.57  & 93.55    & - \\
$^\dagger$ESM-1b~\citep{rives2019esm} & 0.708 & 82.22 & 73.57 & 0.859 & 0.661 & 0.320 & 0.392 & 80.33 & 92.83 & - \\
$^\dagger$MIF-ST~\citep{yang2022masked} & 0.694 & 75.54 & 75.08 & 0.803 & 0.627 & 0.239 & 0.248 & 78.96 & 91.76 & - \\

\midrule
Masked-LM (ESM2-650M)   & 0.691   & 84.78   & 71.88     & 0.866       & 0.676 & 0.344   & 0.402  & 83.68   & 92.28    & 0.80 \\
\ARLM (650M)   & 0.638   & 68.48   & 61.16     & 0.691       & 0.566 & 0.258   & 0.287  & 68.53   & 88.31    & - \\

\midrule
\chl \method (150M) & \chl 0.687          &  \chl 80.98           & \chl {72.17}                     & \chl {0.822}      & \chl 0.662 & \chl 0.328 & \chl0.379 & \chl {82.41} & \chl {92.63} &  \chl -
 \\

\chl \method (650M) & \chl {0.695}           &  \chl {86.41}     & \chl {75.15}                     & \chl {0.875}      & \chl {0.680} & \chl {0.357} & \chl {0.409} & \chl {84.56} & \chl {93.09}  & \chl 0.82 
 \\

\chl \method (3B) & \chl \textbf{0.704}           &  \chl \textbf{90.00}            & \chl \textbf{75.94}                     & \chl \textbf{0.883}      & \chl \textbf{0.687} & \chl \textbf{0.369} & \chl \textbf{0.463} & \chl \textbf{85.32} & \chl \textbf{93.93}  & \chl - \\

\bottomrule
\end{tabular}
}
\vspace{-3mm}
\end{table*}

\subsection{Evaluation of Unconditional Generation}
\label{sec:uncond}

\Figref{fig:uncond_sample} shows the results of \method for unconditional generation, where we evaluate the performance regarding a set of lengths $[100, 200, ..., 900, 1000]$ in intervals of 100.
The reverse process of \method for sampling iterates for 500 steps.
Meanwhile, we also randomly pick the natural sequences of the same length from UniRef50 as reference (denoted as UR50)
We highlight our primary findings as follows:



\paragraph{(1) On Foldability:} 
\ul{DPLM is capable of generating protein sequences with reasonable predicted structures.}
We examine the structural plausibility or foldability of protein sequences using the state-of-the-art single-sequence structure prediction model, \ie, ESMFold~\citep{lin2022esmfold}, and measured by the predicted local distance difference test (\metric{pLDDT}) score, which is considered high confidence if $\metric{pLDDT} > 70$. 
We can find that protein sequences generated by \method achieve the highest \metric{pLDDT} score across all lengths (\Figref{fig:uncond_sample}A). 
Plus, secondary structure analysis of the sequences generated by \method reveals a higher proportion of beta-strands (\Figref{fig:uncond_sample}D), and overall similar to the statistics of known protein structures in Protein Data Bank~\citep[PDB;][]{berman2000protein}.
Moreover, we can see that scaling \method leads to better foldability performance, especially for very long proteins (\Figref{fig:uncond_sample}E).


\paragraph{(2) On Novelty.}
We investigate whether \method can sample sequences possessing novel structures, where we compare the structural similarity against known structures in PDB with \metric{TMScore}.
The highest TMscore is used to measure the novelty of each sequence, which we refer to as \metric{pdb-TM} score.
Overall, \method has relatively higher \metric{pdbTM} than EvoDiff and natural sequences, as shown in \Figref{fig:uncond_sample}B. 
Interestingly, the \metric{pdbTM} score of \method will decrease as protein gets longer than 300 while maintaining the $\metric{pLDDT} > 75$. 
This indicates that \method possesses the ability to sample sequences with structures not similar to PDB across various lengths, with the discrepancy becoming increasingly apparent as the sequence length extends.

\paragraph{(3) On Diversity.}
We quantify the diversity of sequences sampled by \method by \metric{inner-TM} score. 
Specifically, for each sampled candidate, we use ESMFold to predict its structure and compute \metric{TMscore} against the rest.
The average \metric{TMscore} is considered as the diversity.
As shown in \Figref{fig:uncond_sample}C, \method has a considerably low average \metric{inner-TM}, demonstrating that the \method can synthesize structurally diverse sequences.

\paragraph{(4) On Learning:}
\ul{Discrete diffusion is the best-suited probabilistic framework for protein sequence generation, compared to Masked-LM and \textsc{Ar-LM}.}
As shown in \Figref{fig:uncond_sample}F, \method outperforms \MLM and \ARLM in terms of foldability, verifying our motivation to pursue a diffusion protein LM that diffusion is a more proper probabilistic framework for protein modeling.
Moreover, \ARLM also falls short of precisely controlling the length of sampled sequences, making it less flexible in practice.
As revealed in \Figref{fig:uncond_sample}G, we find that despite attaining improved generation quality over ESM2 with directly pre-training \method from scratch (\method-FS), it can bring additional learning challenges and training overheads. 
As such, we leverage a 2-stage training strategy, which consists of masked language modeling as the first stage objective, followed by diffusion objective, solving this problem and obtaining high-quality generation with \metric{pLDDT} closely approaching 90.

\paragraph{(5) Case Study.}
In \Figref{fig:uncond_sample}H, we showcase proteins sampled by \method across various lengths, ranging from 100 to 1000, while more cases are presented in the Appendix.
As the protein gets longer, the complexity of its structure will increase, containing rich helices and sheets.
We also find that \method can sample proteins composed of tandem repeats such as beta-barrel or Kelch repeat domain.





\begin{figure*}[t!]
    \centering
    \vspace{-1mm}
    \includegraphics[width=0.96\linewidth]{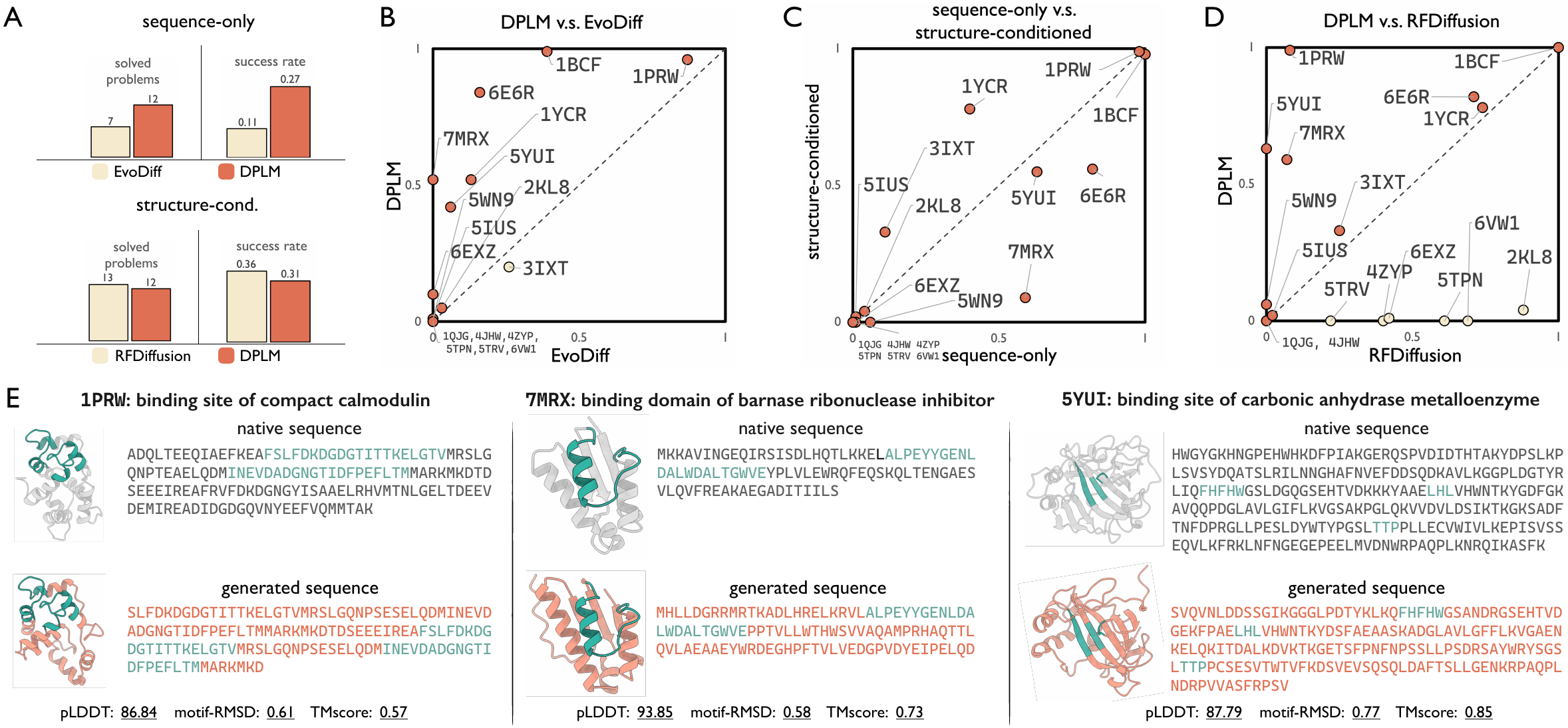}
    \vspace{-4mm}
    \caption{{\sl Evaluation of motif-scaffolding.}
    \textbf{(A)} comparison regarding overall success rate and number of solved problems;
    \textbf{(B)} comparison between sequence-only approaches (\method \vs EvoDiff);  
    \textbf{(C)} comparison between sequence-only \vs structure-conditioned \method; and
    \textbf{(D)} comparison between \method (structure-conditioned and sequence-only \method) \vs RFDffusion;
    \textbf{(E)} case study for three problems.
    }
    \label{fig:evodiff_difflm_motif}
    \vspace{-3mm}
\end{figure*}

\subsection{Evaluation of Protein Representation Learning on Downstream Predictive Tasks}
\label{sec:understanding}

We evaluate \method across a variety of protein predictive tasks~ \citep{su2023saprot,dallago2021flip,xu2022peer}, including protein function prediction (Thermostability and Metal Ion Binding), protein localization prediction (DeepLoc), protein annotation prediction (EC and GO), protein-protein interaction prediction (HumanPPI), where we perform full-parameters supervised fine-tuning on each dataset.
We also include linear probing for secondary structure from TAPE~\citep{rao2019evaluating}.


\paragraph{\method is a superior protein sequence representation learner.}
As demonstrated in \Tabref{tab:results_understanding}, \method outperforms ESM2 across all tasks.
This improved performance is due to the proposed diffusion pre-training, which requires \method to adeptly learn to reconstruct the native sequence from a varied proportion of masking, including very high noise level, in contrast to ESM2 of a fixed 15\% masking ratio.
Under this circumstance, it becomes a much more challenging missing amino acid reconstruction task encouraging the model to capture the deep dependencies from the very context.
Besides, we surprisingly find that \method also closely approaches the performance of SaProt~\citep{su2023saprot}, which is a structure-aware LM that incorporates explicitly protein structures based on Foldseek~\citep{van2023foldseek} and folding models like AlphaFold~\citep{jumper2021AF2}.
This implies that \method may implicitly learn the protein structures from massive sequence data. 
Integrating explicit structural information into \method like \citet{su2023saprot} may bring further benefits, which deserve further exploration.
Our results substantiate our initial premise that \method gains a deeper understanding of protein through the generative learning process, \ie, it learns to better understand proteins by learning to generate them, leading to improved predictive performance.


\subsection{Evaluation of Conditional Generation}
\label{sec:exp_cond}

\subsubsection{Sequence-cond.: Motif-scaffolding}
\label{sec:motif}

The goal of motif-scaffolding requires a valid scaffold to maintain the structure of the given motif such that the original function can be preserved.
Here, we follow the experimental setting in~\citet{alamdari2023protein}, where we (1) initially determine the length of a scaffold and fill the scaffold positions with the mask token; then (2) keep the motif fragment fixed during inference, and sample scaffold conditioned on the motif; and finally use OmegaFold~\citep{wu2022high} to predict the structure of the sampled sequences.
A scaffold is considered successful when it meets two conditions: (1) the \metric{RMSD} between the predicted motif structure and the ground truth, referred to as $\metric{motif-RMSD} < 1\AA$; and (2) the structure should have an overall $\metric{pLDDT} > 70$. 
Overall, we examine 17 motif-scaffolding problems, and for each problem, we sample 100 sequences and then calculate the success rate according to the above criterion.

\paragraph{\method can generate reasonable scaffolds for the given functional motifs.}
As shown in \Figref{fig:evodiff_difflm_motif}, we find that \method outperforms EvoDiff in terms of the number of solved problems and the average success rate. 
Moreover, on the problems that both \method and EvoDiff can solve, the success rate of \method is higher than EvoDiff, except \texttt{3ixt}. 
This indicates that \method excels in motif-scaffolding, preserving the motif structure during scaffold generation.
To gain more insights, we compare \method with structure conditioning (see \Secref{sec:IF}) with state-of-the-art structure designer RFDiffusion~\citep{watson2023RFdiffusion}. We find that \method shows better results in 6 problems, especially for \texttt{1PRW} and \texttt{5YUI}).
We find that utilizing motif structure helps \method make a further improvement on 4 problems compared to the original sequence-only \method, while decreasing performance on the other 6 problems. This implies that for some specific motifs, scaffolding in sequence space may be better. 

The detailed analysis unveiled a common biological property among the motifs observed in these two cases.
Specifically, the motif sequence displayed a remarkable level of evolutionary conservation, playing pivotal roles in binding critical signal passengers (\metric{1PRW}: calmodulin EF hand for calcium binding and \metric{5YUI}: carbonic anhydrase II for CO2 binding). 
Notably, the motif structures predominantly comprised flexible loops. 
Conversely, \metric{5TPN}, \metric{6VW1}, and \metric{2KL8}, which exhibited a distinct advantage in motif scaffolding as indicated by the RFdiffusion, featured rigid helical structures that lacked functional evolutionary conservations. This intriguing phenomenon suggests that DPLM holds great promise as a superior method for constructing structurally flexible yet evolutionarily conserved functional motif scaffolding.





\begin{table}[t!]
   \centering
   \small
   \setlength{\tabcolsep}{2pt}
   \vspace{-2mm}
   \caption{{\sl Performance comparison between \method and different baseline approaches on CATH 4.2 and CATH 4.3 datasets.}
   {\method's results are obtained by \texttt{argmax} decoding (\ie, no sampling).}
   $\dagger$: benchmarked results are quoted from \citet{gao2022pifold}.
   }
   \label{tab:results_cath}

   \resizebox{\linewidth}{!}{%
   \begin{tabular}{llrccc}
   \toprule
   & \multirow{2}{*}{ Models} 
   & \multirow{2}{*}{ Trainable} 
   & \multirow{2}{*}{\metric{AAR}} 
   & \multicolumn{2}{c}{ struct. eval.} \\
    \cmidrule[0.3pt](lr){5-6} & & Params. & & \metric{scTM}  & \metric{pLDDT} \\
   \midrule

   \multirowcell{7}{\rotatebox[origin=c]{90}{CATH 4.2}}
   & $^\dagger$StructTrans~\smallcitep{ingraham2019generative}                     & 1.6M/1.6M  & 35.82&   -   & - \\
   & $^\dagger$GVP~\smallcitep{jing2020gvp}                                       & 1.0M/1.0M  &  39.47 &   - & -  \\
   & $^\dagger$ProteinMPNN~\smallcitep{dauparas2022proteinmpnn}                   & 1.9M/1.9M  &  45.96 &  -  & -    \\
   & PiFold~\smallcitep{gao2022pifold} \                                        & 6.6M/6.6M  &  {51.66} &   {-} &{-}  \\ 
  \cmidrule[0.5pt](lr){2-6}
  & ProteinMPNN + CMLM                                                 & 1.9M/1.9M  &48.62     & 0.87 & 74.07 \\ 

  &  \textsc{LM-Design} (\textit{w/} ProtMPNN encoder)              &  5.0M/650M   &  {54.41} &  0.88 &  77.07 \\
  & \chl \method (\textit{w/} ProtMPNN encoder)              & \chl 5.0M/650M   & \chl \textbf{54.54} & \chl 0.88 & \chl \textbf{77.12} \\
  \midrule

  \multirowcell{6}{\rotatebox[origin=c]{90}{CATH 4.3}}
  & PiFold~\smallcitep{gao2022pifold} & 6.6M/6.6M    &  {51.66} & - & - \\ 
  & GVP-Transformer~\citep{hsu2022esmif}                                 & 142M/142M  &  51.60 &   - & -  \\
  &  \textsc{LM-Design} (\textit{w/} GVP-Trans encoder)          &  6.3M/650M  & 56.49 &  0.85 & 74.89  \\

  \cmidrule[0.5pt](lr){2-6}
  & \chl \method-150M (\textit{w/} GVPTrans encoder)                                               & \chl 3.1M/{150M}   &\chl {53.27}   & \chl {0.85}  &\chl \textbf{75.31}  \\ 
  & \chl \method-650M (\textit{w/} GVPTrans encoder)                                               & \chl 6.3M/{650M}   &\chl {56.61}   & \chl {0.86}  &\chl {76.78}  \\ 
   & \chl \method-3B (\textit{w/} GVPTrans encoder)                                               & \chl 68.2M/{3.0B}   &\chl \textbf{59.44}   & \chl \textbf{0.86}  &\chl \textbf{77.12}  \\ 
\bottomrule
\end{tabular}
}
\end{table}

\subsubsection{Structure-conditioned: Inverse Folding}
\label{sec:IF}
The goal of inverse folding is to find an amino acid sequence that can fold to a given protein backbone structure. 
We follow \textsc{LM-Design}~\citep{zheng2023structure} to implant a structural adapter into the last network layer of \method, and use GVP-Transformer Encoder~\citep{hsu2022esmif} as the expert protein backbone structure encoder.
We assess \method on CATH 4.2 and 4.3~\citep{orengo1997cath}.
We use amino acid recovery (\metric{AAR}) for sequence evaluation, whilst for structure evaluation, we first predict the structure of the generated sequence using ESMFold, then calculate the \metric{pLDDT} score and self-consistency TM-score (\metric{scTM}) between predicted structure and the input one.

\paragraph{\method yields sequences that can confidently fold into the given backbone structure.}
As shown in \Tabref{tab:results_cath}, \method can outperform or be on par with our strong baselines, including the state-of-the-art approach~\textsc{LM-Design}~\citep{zheng2023structure}, manifesting in \metric{AAR}, and most importantly, decent performance regarding structure evaluation ($\metric{scTM} =0.85$ and $\metric{pLDDT} > 76$).
We suggest this derives from the well-learned protein sequence knowledge of \method.
When given structure backbone information, \method can leverage this advantage and generate the sequence whose structure is both plausible and similar to the reference.

\begin{figure}[t!]
    \centering
    \includegraphics[width=\linewidth]{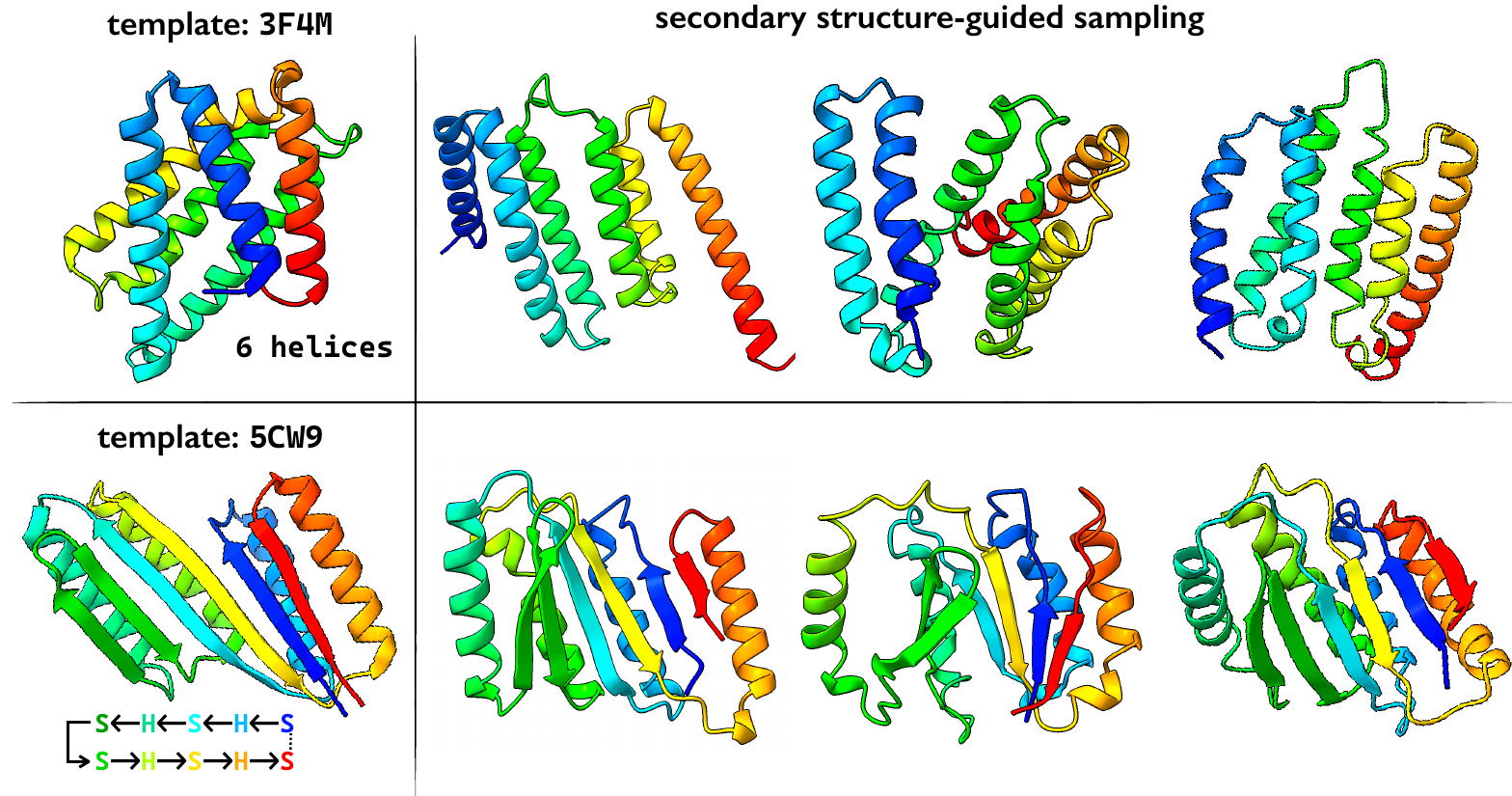}
    \vspace{-8mm}
    \caption{{\sl Secondary structure guided conditional sampling.}
    The first case contains 6 alpha-helices, The second case is much more complicated as a globally twisted structure with interleaved alpha-helices and beta-strands, where the N-terminus and C-terminus are structurally contiguous.}
    \label{fig:guided_sample}
\end{figure}

\subsubsection{Controllable Generation: Secondary Structure Guided Protein Sampling}
\label{sec:controllable}
Classifier guidance is preferred for its flexible control over the generation process without retraining for each new condition, especially beneficial in scenarios with too limited labeled data to directly attain conditional models.
Here we showcase how to guide \method to generate proteins satisfying desired secondary structures.
We train a secondary structure prediction (SSP) model as a sequence labeling task on TAPE dataset. 
We then integrate this SSP discriminative model into \method to provide guiding signals.

\paragraph{\method enjoys plug-and-play programmability.}
\Figref{fig:guided_sample} showcases that the proposed discrete classifier guidance helps steer a pre-trained \method to generate samples satisfying provided secondary structure annotations extracted from template natural proteins.
These findings suggest that \method is highly programmable, and its full potential of generative capabilities can be realized in a plug-and-play fashion, indicating that \method preserves the appealing characteristic of controllable generation inherent in diffusion models, but for discrete data.
This flexibility to swiftly adapt to the evolving needs of users across a broad spectrum of preferences is also significant in practical applications with time and computational paramount.




\section{Discussions}
\label{sec:conclusion}

In this paper, we introduce diffusion protein LM (\method), a versatile protein LM that is capable of both protein sequence generation and representation learning. 
We further develop several conditioning strategies for various needs of conditional generation, including sequence conditioning, cross-modal conditioning, and programmable generation with plug-and-play discrete classifier guidance.

Despite these promising results, there remain several limitations and future work directions deserving to be explored.

\begin{compactitem}

\item[(i)] \textit{Exploring \method's conditional generation for wider applications.}
We can further extend the cross-modal conditioning strategy of \method to more diverse modalities as conditioners, including MSA-conditioned homologous sequence generation, small molecule-conditioned binder design for ligands, antigen-conditioned antibody CDR design, among others. 
Also, the inclusion of demonstrations featuring plug-and-play classifier-guided controllable generation is essential for more scenarios toward diverse user preferences, \eg, structural symmetry, superfamily, binding affinity, thermostability, fluorescence, and beyond.

\item[(ii)] \textit{\method can further benefit from best practices of cutting-edge technical advancement in the vastness of large language models (LLMs).}
For example, \textbf{(1)} long context extension~\citep{chen2023extending} can rapidly adapt \method to handle very long proteins beyond its training length limit, and offering potential for modeling exceptionally long biological sequences such as DNAs and RNAs, unifying and deciphering the languages associated with the central dogma of life; 
\textbf{(2)} fine-tuning \method with human feedback or even wet-lab experimental feedback, leveraging reinforcement learning~\citep[RL;][]{ouyang2022instructgpt}, direct preference optimization~\citep[DPO;][]{rafailov2024dpo}, and self-play fine-tuning~\citep{chen2024spin};
\textbf{(3)} eliciting instruction-following and in-context learning~\citep{wei2022emergent} analogs for protein LMs can also be a promising direction would fully harness \method's learned knowledge.

\item[(iii)] \textit{It is imperative to integrate protein structure modeling into \method.}
The advance of protein structure modeling manifest tremendous success, including AlphaFold~\citep{jumper2021AF2}, ESMFold~\citep{lin2022esmfold} for structure prediction, RFDIffusion~\citep{watson2023RFdiffusion}, Chroma~\citep{ingraham2023chroma} for structure design, and even full-atom molecular modeling, \eg, the latest generation of AlphaFold~\citep{deepmind2023AF3} and RF-AA~\citep{krishna2023RFAA}. 
Developing a universal protein language model with the next-generation \method, which accounts for both sequence and structure, is a particularly promising avenue.

\end{compactitem}

We leave these exciting directions as future work.

\section*{Impact Statement}
Our work on protein generation and representation learning can be used in developing potent therapeutic macromolecues such as antibodies and accelerate the research process of drug discovery. 
Our method may be adapted to other scenarios of computer-aided design, such as small molecule design, material design, and chip design. 
It is also needed to ensure the responsible use of our method and refrain from using it for harmful purposes.

\section*{Acknowledgements}
We thank anonymous reviewers for their insightful feedback. 
We would like to especially thank Dr. Hang Li for insightful discussions on the project and feedback on the manuscript that help shape this study. 
We thank Yi Zhou, Jing Yuan, Dr. Yilai Li and Jiasheng Ye for their valuable comments.

\bibliography{references}
\bibliographystyle{icml2024}

\newpage
\clearpage
\appendix
\section{Reparameterizaed Discrete Diffusion Models (RDM)}
\label{app: rdm}
\method uses reparameterized discrete diffusion model (RDM) as its discrete diffusion framework~\citep{zheng2023reparameterized}.
Here we briefly summarize its basic training and sampling. Please refer to~\citet{zheng2023reparameterized} for more details.

\citet{zheng2023reparameterized} shows that the backward transition of discrete diffusion models $q(\bm{x}^{(t-1)}|\bm{x}^{(t)}, \bm{x}^{(0)})$ can be rewritten as 
\begin{align}
   & q(\bm{x}^{(t-1)}|\bm{x}^{(t)}, \bm{x}^{(0)}) \nonumber \\
    &= \begin{cases}
\lambda_{t-1}^{(1)}\bm{x}^{(t)} + (1-\lambda_{t-1}^{(1)})\bm{q}_{\text{noise}}, & \text{if } \bm{x}^{(t)}=\bm{x}^{(0)} \nonumber \\
\lambda_{t-1}^{(2)}\bm{x}^{(0)} + (1-\lambda_{t-1}^{(2)})\bm{q}_{\text{noise}}(\bm{x}^{(t)}), 
 & \text{if } \bm{x}^{(t)} \not=\bm{x}^{(0)}
\end{cases}
\label{eqn: backward transition}
\end{align}
where $\bm{q}_{\text{noise}}(\bm{x}^{(t)}) = \beta_t\bm{x}^{(t)} + (1-\beta_t)\bm{q}_{\text{noise}}$, and both $\lambda_{t-1}^{(1)}$ and $\lambda_{t-1}^{(2)}$ are constants relating to $\beta_t$ and $\beta_{t-1}$. This reformulation interprets the backward transition as a mixture distribution. Sampling from it is equivalent to first sampling from a Bernoulli distribution and then the corresponding component distribution, \ie,

$$
\begin{aligned}
v_{t-1}^{(1)}\sim \text{Bernoulli}\left(\lambda_{t-1}^{(1)}\right) , &
\bm{u}_t^{(1)}\sim \texttt{Cat}\left(\bm{u}; \bm{p}=\bm{q}_{\text{noise}}\right),&\\
v_{t-1}^{(2)}\sim \text{Bernoulli}\left(\lambda_{t-1}^{(2)}\right) , &
\bm{u}_t^{(2)}\sim \texttt{Cat}\left(\bm{u}; \bm{p}=\bm{q}_{\text{noise}}(\bm{x}_t) \right),&
\end{aligned}\\
$$
$$
\bm{x}^{(t-1)}=\left\{
\begin{aligned}
v_{t-1}^{(1)}\bm{x}^{(t)} + \left(1-v_{t-1}^{(1)}\right)\bm{u}_t^{(1)},&&\text{if }\bm{x}^{(t)}=\bm{x}^{(0)}\\
v_{t-1}^{(2)}\bm{x}^{(0)} + \left(1-v_{t-1}^{(2)}\right)\bm{u}_t^{(2)},&&\text{if }\bm{x}^{(t)}\not=\bm{x}^{(0)}
\end{aligned}
\right..
$$

This reparameterizes the transitions $q(\bm{x}^{(t-1)}|\bm{x}^{(t)}, \bm{x}^{(0)})$ and $p_{\theta}(\bm{x}^{(t-1)}|\bm{x}^{(t)})$ into $q(\bm{x}^{(t-1)},\bm{v}^{(t-1)}|\bm{x}^{(t)}, \bm{x}^{(0)})$ and $p_{\theta}(\bm{x}^{(t-1)},\bm{v}^{(t-1)}|\bm{x}^{(t)})$. 
With this reparameterization, the training objective of diffusion models (\ie, the variational bound of negative log-likelihood) becomes
\begin{align}
& -\mathbb{E}_q(\bm{x}_{1:T}, \bm{v}_{1:T}|\bm{x}_0)\left[\log \frac{p_\theta(\bm{x}_0,\bm{x}_{1:T},\bm{v}_{1:T})}{q(\bm{x}_{1:T},\bm{v}_{1:T}|\bm{x}_0)} \right] \nonumber \\
& =\mathcal{J}_1 + \sum_{t=2}^T \mathcal{J}_t + \text{const.}, \nonumber
\end{align}
where $\mathcal{J}_1=-\mathbb{E}_{q(\bm{x}_1|\bm{x}_0)}\left[\log p_\theta(\bm{x}_0|\bm{x}_1)\right]$ and \citet{zheng2023reparameterized} shows that $\mathcal{J}_t$ can be simplified into a weighted cross-entropy loss. 
Since each token is modeled conditionally independently, so we can consider the backward transition for each token, and sum the losses for them. For i-th position, the backward transition is $q(\bm{x}_i^{(t-1)},\bm{v}_i^{(t-1)}|\bm{x}_i^{(t)}, \bm{x}_i^{(0)})$. As shown in \citet{zheng2023reparameterized} appendix C, the loss at i-th token can be written as 
\[
\small
\begin{aligned}
   &\mathcal{J}_{t,i} = \mathbb{E}_{q(\bm{v}_i^{(t-1)})} \nonumber \\
   & \left[ 
   \text{KL}[
   q(\bm{x}_i^{(t-1)}|\bm{v}_i^{(t-1)},\bm{x}_i^{(t)}, \bm{x}_i^{(0)})||  
  p_{\theta}(\bm{x}_i^{(t-1)}|\bm{v}_i^{(t-1)},\bm{x}_i^{(t)})] \right] \nonumber
\end{aligned}
\]
Let $b_i{(t)}=\mathbf{1}_{x_i^{(t)} \neq x_i^{(0)}}$,
$q(\bm{x}_i^{(t-1)}|\bm{v}_i^{(t-1)},\bm{x}_i^{(t)}, \bm{x}_i^{(0)})$ can be written as:
\begin{align}
   & q(\bm{x}_i^{(t-1)}|\bm{v}_i^{(t-1)},\bm{x}_i^{(t)}, \bm{x}_i^{(0)}) \nonumber \\
    &= \begin{cases}
v_{t-1,i}^{(1)}\bm{x}_i^{(t)} + (1-v_{t-1,i}^{(1)})\bm{q}_{\text{noise}}  &\text{if } b_i{(t)}=0, \nonumber \\
v_{t-1,i}^{(2)}\bm{x}_i^{(0)} + (1-v_{t-1,i}^{(2)})\bm{q}_{\text{noise}}  &\text{if } b_i{(t)}=1,
\end{cases}
\end{align}
And $p_{\theta}(\bm{x}_i^{(t-1)}|\bm{v}_i^{(t-1)},\bm{x}_i^{(t)})$ can be written as:
\begin{align}
   & p_{\theta}(\bm{x}_i^{(t-1)}|\bm{v}_i^{(t-1)},\bm{x}_i^{(t)}) \nonumber \\
    &= \begin{cases}
v_{t-1,i}^{(1)}\bm{x}_i^{(t)} + (1-v_{t-1,i}^{(1)})\bm{q}_{\text{noise}}  &\text{if } b_i{(t)}=0, \nonumber \\
v_{t-1,i}^{(2)}p_{\theta}(\bm{x}_i^{(0)}|\bm{x}^{(t)}) + (1-v_{t-1,i}^{(2)})\bm{q}_{\text{noise}}  &\text{if } b_i{(t)}=1,
\end{cases}
\end{align}
Therefore, the loss at i-th token can be computed by enumerating all cases with respect to $\bm{v}_i^{(t-1)}$ and $b_i(t)$. 
As noted in \citet{zheng2023reparameterized}, the KL divergence is equal to $-\log p_{\theta}(x_i^{(0)}|x^{(t)})$ when $v_{t-1,i}^{(2)}=1$ and $b_i(t)=1$, while in other cases the KL divergence is 0. 
So we have:
\[
\small
\begin{aligned}
   & \mathcal{J}_t = \sum_{1 \leq i \leq L}\mathcal{J}_{t,i} \nonumber \\
   &= \sum_{1 \leq i \leq L} \mathbb{E}_{q(\bm{v}_i^{(t-1)})} \bigg[ \\
   & \text{KL}[q(\bm{x}_i^{(t-1)}|\bm{v}_i^{(t-1)},\bm{x}_i^{(t)}, \bm{x}_i^{(0)})||p_{\theta}(\bm{x}_i^{(t-1)}|\bm{v}_i^{(t-1)},\bm{x}_i^{(t)})]\bigg] \nonumber \\
    &= 
\sum_{1 \leq i \leq L} q(\bm{v}_i^{(t-1)}=1)\cdot b_i(t) \cdot (-\log p_{\theta}(x_i^{(0)}|x^{(t)})) \nonumber \\
&= 
-\lambda^{(t-1)}  {\sum_{1 \leq i \leq L}} b_i(t) \cdot \log p_{\theta}(\bm{x}_i^{(0)}|\bm{x}^{(t)}) \nonumber
\end{aligned}
\]
Notably, training with different noise schedules only differs in the weighting of the objective.

During sampling, RDM leverages this observation and proposes to employ a discriminative approach.
Specifically, it denoises a token only when it receives a top-$k$ score (log-probability) from the network where $k$ in each step is determined by a denoising schedule. The overall sampling process is shown in algorithm \ref{alg:rdm:sampling}.

\begin{algorithm}[h] %
   \caption{Sampling from RDM}
   \label{alg:rdm:sampling}
    \begin{algorithmic}
    \State {\bfseries Input:} trained network $f_\theta\left(\cdot\right)$ and temperature $\tau$.
    \State {\bfseries Output:} generated sample $\xt{0}$.
    \For{$n = 1,2,\dots,N$}
        \State Initialize $\bm{x}_{T,n} \sim q_{\text{noise}}$;
        \State Initialize $b_{T,n} = 0$;
    \EndFor
    \For{$t = T,\dots,1$}
        \For{$n = 1,2,\dots,N$}
            \State Draw $\widetilde{\bm{x}}_{0,n} \sim \operatorname{Categorical}\left(f_\theta\left(\bm{x}_{t,n}\right)\!/\tau\right)$;
            \State Generate $\bm{v}_{t-1,n}$ according to $\log p(\widetilde{\bm{x}}_{0,n})$
            \If{$b_{t,n} = 1$}
            \State Draw $\bm{u}_{t,n}^{(1)} \sim q_\text{noise}$;
            \State $\bm{x}_{t-1,n} = v_{t-1,n}^{(1)}\bm{x}_{t,n} + \left(1-v_{t-1,n}^{(1)}\right)\bm{u}_{t,n}^{(1)}$;
            \Else
            \State Draw $\bm{u}_{t,n}^{(2)} \sim q_\text{noise}(\bm{x}_{t,n})$;
            \State $\bm{x}_{t-1,n} = v_{t-1,n}^{(2)}\widetilde{\bm{x}}_{0,n} + \left(1-v_{t-1,n}^{(2)}\right)\bm{x}_{t,n}^{(2)}$;
            \EndIf
            \State Let $b_{t-1,n} = b_{t,n} \land v_{t-1,n}^{(1)} \lor v_{t-1,n}^{(2)}$;
        \EndFor
    \EndFor
    \State {\bfseries Return} $\bm{x}_{0,1:N}$.
    \end{algorithmic}
\end{algorithm}


\section{Training Stratety}
\subsection{Pre-training of \method}
During the training phase, we investigate two different approaches: (1) training with the diffusion objective from scratch, and (2) what we refer to as two-stage training, which consists of initial training with the masked language modeling (MLM) objective followed by continuous training with the diffusion objective.

\paragraph{Empirical Observation.} In our preliminary experiments, we observed that discrete diffusion pre-training "from scratch (FS)" often yielded instability in the form of frequent loss spiking, therefore hurting our model performance.

\paragraph{Our Hypothesis.} We noticed that the absorbing diffusion objective leads to a variable masking ratio ranging from 0\% to 100\%, while conventional MLM objective's masking ratio keep fixed at 15\% such that a masked LM is always exposed to rich condition of 85\% observation/context. 
In other words, in contrast to MLM, for absorbing discrete diffusion models like DPLM, in some of the extreme cases there are nearly all tokens getting masked, which means that the model is required to recover all tokens of the ground-truth sequence from nothing). 
This could impose severe learning challenges, especially at the early phase of pre-training, where the model has yet not acquired sufficient capability to extract informative features and correlations from limited observation.

\paragraph{Solution in Principle.}  Inspired by the success of curriculum learning in non-autoregressive text generation~\citep{qian2020glancing,gu2021fully, guo2020fine, guo2020incorporating,wang2022xlm}, we suggested that a masking warmup strategy could mitigate this issue, where we can start with a small upper bound of masking ratio (e.g., 15\% as conventional MLM, to preserve a high proportion of observation) in the early phase of pre-training, and then gradually increase the masking ratio towards the authentic discrete diffusion objective during pre-training.

\paragraph{Solution in Practice.} 
In the current form of our manuscript, adhering to this principle, we proposed a two-stage training method: initialized DPLM from an established masked LM, either from our in-house pre-trained one or from the official ESM-2 checkpoint, and then trained DPLM by the discrete diffusion language modeling objective afterwards. Though it may or may not lead to the best model performance, it offers the possibilities of standing on the shoulders of any pre-trained masked LM such as ESM2 or the advanced LM architecture such as a Llama/Mistral-style masked LMs, in the broad open-source AI community. This also enables us to bypass the time-consuming process of gradual masking warmup during pre-training. As a result, this can be the most efficient and effective approach in practice, which also shares a similar principle as finetuning from RosettaFold in RFDiffusion~\citep{watson2023RFdiffusion}.

Although discrete diffusion pre-training from scratch is challenging, we have also explored several ways to improve it. 
As shown in the \Figref{fig:uncond_sample}G, DPLM can achieve comparable performance through pre-training from scratch. 
More specifically, we found that (1) gradient norm clipping can effectively help stabilize training process of discrete diffusion language modeling, greatly reducing the chance of loss spiking and gradient nan. 
In addition, we also found that (2) training longer (i.e. scaling compute) is another key to attain a good ultimate model performance (trained for 300k steps). 

We are also curious about the performance of the vanilla masking warmup strategy and would like to see if it can lead to a better pre-trained DPLM. The training dynamics of discrete diffusion based DPLM is an interesting and exciting direction deserving further exploration, and we leave these as our future work.


\subsection{Pre-training of \ARLM baseline}

We pretrain a \ARLM using autoregressive training objective. 
In order to be comparable with \method, the autoregressive language model we trained adopts the same architecture as \method. 
To be capable of adapting the autoregressive training, we modify the mask matrix of the attention module to causal mask, which guarantees each token can only attend the previous position and keep unseen for future. 
The training objective is next word prediction, and we process the input sequence with teacher forcing for efficient parallel training. 
During decoding, we start with \metric{<bos>} token, sample one token each timestep from left to right, and the sampled token in the current timestep will be concatenated to the end of the sequence, becoming input for next timestep. 
The decoding process terminates until \metric{<eos>} token is sampled. Because we can not know when to obtain the \metric{<eos>} token in advance, we can not decide the length of sampled sequence. 
We attempt to force the sampling length by modifying the sampling probability: the probability of \metric{<eos>} is 1 when and only when the sequence length is up to the predefined length, while 0 in all the previous timesteps. 
However, we observe this will decline the quality of sampled sequence significantly.

\section{Reasons for choosing absorbing discrete diffusion}

We employed absorbing discrete diffusion as the pre-training method, instead of other forms of discrete diffusion or latent diffusion, for the following considerations.
\subsection{Regarding discrete diffusion (DD) definitions (multinomial vs absorbing as proposed in D3PM)}

\paragraph{Intuition.} We favor absorbing DD since the learning objective of absorbing DD generalizes existing language modeling objectives, as highlighted in section 4 of \citet{austin2021structured}, In particular, absorbing DD is a natural extension of the masked language modeling (MLM) objective, which has been thoroughly studied~\citep{wettig2023should} in the field of NLP and widely proven to be a robust and effective sequence learning protocol. 
In contrast, multinomial/uniform-DD resembles (tranfitional) denoising autoencoders~\citep{savinov2021step}. 
There is little solid evidence and remains highly unclear about (mutlinomial) denoising autoencoder as a sequence learning objective can scale up w.r.t data, model size and application scenarios.

\begin{figure}[h!]
    \centering
    \vspace{2mm}
    \includegraphics[width=0.9\linewidth]{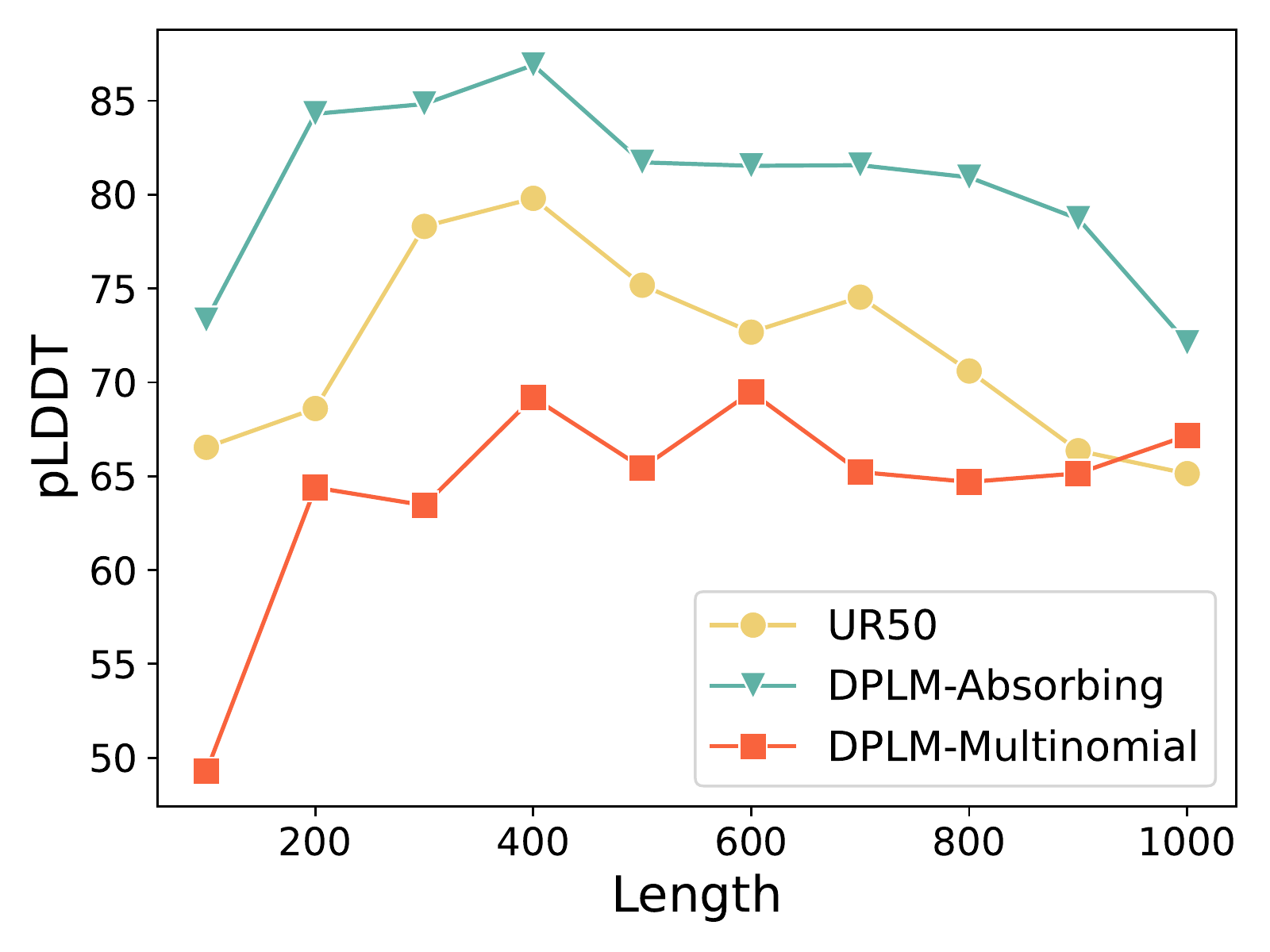}
    \vspace{-4mm}
    \caption{The unconditional sampling performance of multinomial/uniform-DD and absorbing-DD.} 
    \label{fig:multinomial-absorbing}
    \vspace{4mm} 
\end{figure}

\paragraph{Empirical verification.} In our preliminary exploration, we have studied the performance of multinomial/uniform-DD and absorbing-DD. Here we provide the result regarding unconditional sampling, as shown in \Figref{fig:multinomial-absorbing}. 
We can find that DPLM-absorbing generally manifests better performance than DPLM-multinomial across different lengths.

\subsection{Regarding latent diffusion}

We reason that latent diffusion for discrete sequence data, which performs Gaussian diffusion in continuous embedding space, requires additional lossy continuous relaxation of discrete sequence data like protein sequence, which does not fit the discrete nature of protein sequence and is not necessarily the best choice for modeling discrete sequence data. 
Recent study~\citep{meshchaninov2024diffusion} presents a latent diffusion model on protein LM embedding, where the pLDDT score unconditional sampling attains reasonable pLDDT in their paper, while our discrete diffusion based approach still excels.

\section{Additional Experimental Details}
\subsection{A modified unconditional sampling strategy}

The sampling algorithm proposed in the \citet{zheng2023reparameterized} is to unmask positions with top-k prediction score (log-probability) predicted by $p_{\theta}(\xt{0}|\xt{t})$, and mask all the rest position in each denoising step. 
However, we find that if we use this sampling algorithm to sample sequence unconditionally, the sampled sequence will collapse to trivial pattern, such as repeating with a single amino acid.
We suggest this is because, without any additional conditions, the model initially tends to give a higher prediction score to the amino acids that appear frequently in the training set.
Subsequently, based on these high-frequency amino acid tokens, the model will continue to sample the same tokens beside these tokens with high confidence. 
The other amino acid can also be sampled, but possibly with a lower prediction score, thereby leading to be dropped according to the top-k sampling algorithm.
Then, this amino acid will spread throughout the entire sequence like a virus, forming a sequence composed entirely of the same amino acids.

In response, we impose a slight disturbance during sampling, utilizing the Gumbel-Max trick. 
The Gumbel-Max trick is a procedure for drawing a sample from a categorical distribution using Gumbel-distributed random variables. Let's assume we have a discrete random variable $X$ with distribution $p_\theta(\bm{x} = i) = p_i$ for $i=1,\ldots,K$. Now, consider the variables $g_i = -\log{(-\log{U_i}})$ where $U_i$ is a variable uniformly distributed on $(0,1]$. The $g_i$ are random variables following a Gumbel distribution. The key to the Gumbel-Max trick is this relationship:
\begin{align}
i^* = \argmax_{i}  \{ \Tilde{p}_i \}, ~\text{where}~ \Tilde{p} \propto \exp{g_i + \log {p_i}}
\end{align}
This operation provides a sample from the discrete distribution $p_\theta(\bm{x}=i)$. In other words, the category corresponding to the maximum value is the results of sampling.
But in the other hand, the maximum value, \ie $g_i + \log p_i$, is not equal to the original log-probability, which is actually the prediction score in our sampling algorithm.
Therefore, the Gumbel-Max trick helps us sample an amino acid with a slightly modified prediction score while maintaining the original distribution.
As a result, the previously dominant amino acid with the highest prediction score may be discarded, and a variety of other amino acids may be retained, thereby avoiding falling into a trivial pattern such as repeating with a single amino acid. We find that this technique can significantly reduce the number of trivial cases and further improve the diversity.





\subsection{Delve deeply into the pLDDT score of unconditional sampling}

According to the \Figref{fig:uncond_sample}A, we surprisingly find that the pLDDT score of \method unconditional sampling is even higher than the UniRef50 dataset. 
We investigate this phenomenon as follows.

\paragraph{Regarding the lower pLDDT in UniRef50.} 
The lower pLDDT here is because UniRef50 contains some data with lower structural plausibility, such as sequences with a large number of repetitive patterns. 
These cases decrease the average pLDDT of UniRef50, for instance sequence \metric{ADADAD...ADADAD} with pLDDT 35.14.

We also investigate the average pLDDT of the PDB data where the pLDDT score of PDB is similar to DPLM, suggesting that DPLM learns to generate protein sequences with overall similar structural characteristics as PDB, as shown in \Tabref{tab:plddt_pdb}.

\begin{table}[h]
   \centering
   \small
   \setlength{\tabcolsep}{2pt}
   \vspace{2.5mm}
   \caption{ pLDDT score of UniRef50, PDB and \method unconditional sampling.}
   \label{tab:plddt_pdb}
   \begin{tabular}{cccc}
\toprule
Length & UniRef50 & PDB & \method \\
\midrule
100    & 66.54   & 84.62   & 70.66      \\
200    & 68.61   & 79.32   & 83.55      \\
300    & 78.30   & 84.51   & 82.39      \\
400    & 79.80   & 80.49   & 86.75      \\
500    & 75.17   & 77.79   & 82.56      \\
\midrule
avg. pLDDT & 73.68 & \textbf{81.34}   & 81.18    \\ 
\bottomrule
\end{tabular}
   \vspace{4mm}
\end{table}

\paragraph{Regarding mode collapse.} We also want to investigate whether \method collapses into the modes with high pLDDT sequences.
To verify this, we evaluate the pseudo-perplexity of DPLM against subsets of UR50 sequences of high pLDDT and low pLDDT. 
The results are shown in \Tabref{tab:pseudo-ppl}. 
We can find that the ppl of less-structural proteins (pLDDT $<$ 50) is similar to the structural proteins (pLDDT $>$ 70), suggesting that DPLM equally learns protein sequences with diverse structural patterns.

\begin{table}[h]
   \centering
   \small
   \setlength{\tabcolsep}{2pt}
   \vspace{2.5mm}
   \caption{ pseudo-ppl of less structural and more structural sequences.}
   \label{tab:pseudo-ppl}
   \begin{tabular}{ccc}
\toprule

 & less structural sequences  & more structural sequences \\

\midrule
pseudo-ppl    & 2.36   & 2.55      \\

\bottomrule
\end{tabular}
   \vspace{4mm}
\end{table}

In conclusion, we would like to provide a possible explanation for this phenomenon.
We suggest that from a perspective of probabilistic graphical model (PGM), more structured data is generally more easy to learn due to stronger correlation between its elements. 
As such, we hypothesize the learning dynamics of protein LMs from through evolutionary sequences is first to digest those more structural proteins as co-evolutionary effects between amino acids play a prevailing and vital role in folding patterns, and then start to learn those less structural folding patterns, which could be long-tailed.

This could somehow relate to the so-called emergence phenomenon in the realm of LLMs, where scaling up LLMs leads to "grokking" those long-tailed abilities. 
We would leave a study of the learning dynamics of protein LM as an exciting future investigation and hopefully can bring some interesting insights to the community.

\subsection{Sequence-conditional generation: motif-scaffolding}
\begin{table}[t]
\centering
\vspace{-1.5mm}
\caption{Results of the success rate of each problem, the number of the solved problems and the average success rate across 17 motif-scaffolding problems. Here we follow previous work to use OmegaFold as the folding model.}
\label{tab:motif_success}
\vspace{1.5pt}
\resizebox{\linewidth}{!}{%
\begin{tabular}{ccccc}
\toprule
\multirow{2}{*}{} & \multicolumn{2}{c}{{seq-only}} & \multicolumn{2}{c}{{struct-cond.}} \\ 
\cmidrule[0.5pt](lr){2-3}  \cmidrule[0.5pt](lr){4-5}
                  & EvoDiff        & DPLM                 & RFDiffusion        & DPLM   \\
\midrule
1bcf              & 0.39           & \textbf{0.99}        & 1.00                 & 1.00                    \\
1prw              & 0.87           & \textbf{0.96}        & 0.08               & \textbf{0.99}        \\
1qjg              & 0.00              & 0.00                    & 0.00                  & 0.00                    \\
1ycr              & 0.13           & \textbf{0.52}        & 0.74               & \textbf{0.78}        \\
2kl8              & 0.03              & \textbf{0.05}        & 0.88               & 0.05                 \\
3ixt              & 0.26           & 0.20                  & 0.25               & \textbf{0.33}        \\
4jhw              & 0.00              & 0.00                    & 0.00                  & 0.00                    \\
4zyp              & 0.00              & \textbf{0.01}        & 0.40                & 0.01                 \\
5ius              & 0.00              & \textbf{0.10}         & 0.02               & \textbf{0.10}         \\
5tpn              & 0.00              & 0.00                    & 0.61               & 0.00                    \\
5trv              & 0.00              & 0.00                    & 0.22               & 0.00                    \\
5wn9              & 0.00              & \textbf{0.01}        & 0.00                  & \textbf{0.01}        \\
5yui              & 0.06           & \textbf{0.42}        & 0.00                  & \textbf{0.63}        \\
6e6r              & 0.16           & \textbf{0.84}        & 0.71               & \textbf{0.84}        \\
6exz              & 0.00              & \textbf{0.01}        & 0.42               & 0.01                 \\
6vw1              & 0.00           & 0.00                    & 0.69               & 0.00                    \\
7mrx              & 0.00              & \textbf{0.59}        & 0.07               & \textbf{0.59}        \\ 
\midrule
pass rate         & 7/17           & \textbf{12/27}       & 13/17              & 12/17                \\
avg. success rate & 0.09           & \textbf{0.27}        & 0.36               & 0.31                 \\ 
\bottomrule
\end{tabular}
}
\vspace{4mm}
\end{table}

The overall motif-scaffolding results are shown in \Tabref{tab:motif_success}.
We sample 100 scaffold sequences for each motif scaffolding case, and compute the success rate according to the standard mentioned in section~\ref{sec:motif}. Furthermore, we also show the pass rate (e.g. the number of solved problems) and the average success rate for all problems. 
We use sequence-only and structure-conditioned sampling paradigms. 
For sequence-only sampling, DPLM generates scaffold according to the motif sequence fragment.
For structure-conditioned sampling, DPLM makes generation by leveraging both sequence and structure information of motif. 
Specifically, as noted in section~\ref{sec:IF}, we utilize the pre-trained GVPTransformerEncoder and structural adapter to process the motif structure.
DPLM is able to solve 12 of 17 motif scaffolding problems. 
The overall success rate is 0.27 for sequence-only sampling, while 0.31 for structure-conditioned sampling.
It should be noted that not all problems are suitable for using structure information.
We recommend using structure-conditioned sampling for 1YCR, 1PRW, 3IXT and 5YUI, while sequence-only sampling for others.

\begin{table}[t]
\centering
\vspace{-1.5mm}
\caption{Motif-scaffolding results evaluated by ESMFold. }
\label{tab:motif_esmfold}
\vspace{1.5pt}
\resizebox{0.9\linewidth}{!}{%
\begin{tabular}{cccc}
\toprule
\multirow{2}{*}{} & \multicolumn{2}{c}{{seq-only}} & \multicolumn{1}{c}{{struct-cond.}} \\ 
\cmidrule[0.5pt](lr){2-3} \cmidrule[0.5pt](lr){4-4} 
                  & EvoDiff        & DPLM &                   \multicolumn{1}{c}{DPLM}   \\
\midrule
1bcf              & 0.38           & \textbf{1.00}         & \textbf{1.00}                                     \\
1prw              & 0.36           & \textbf{0.75}        & \textbf{0.81}                       \\
1qjg              & 0.00              & 0.00                     &0.00                                      \\
1ycr              & 0.03           & \textbf{0.27}         & \textbf{0.48}                       \\
2kl8              & 0.00              & \textbf{0.01}        & \textbf{0.01}                              \\
3ixt              & 0.09           & \textbf{0.15}                  &  \textbf{0.37}                      \\
4jhw              & 0.00              & 0.00                    & 0.00                                      \\
4zyp              & 0.00              & 0.00
&  \textbf{0.01}                                 \\
5ius              & 0.00              & 0.00
& 0.00                        \\
5tpn              & 0.00              & 0.00                    & 0.00                                   \\
5trv              & 0.00              & 0.00                    & 0.00                                   \\
5wn9              & 0.00              & 0.00
& 0.00                          \\
5yui              & 0.05           & \textbf{0.94}        &  \textbf{0.94}                          \\
6e6r              & 0.03           & \textbf{0.79}        &  \textbf{0.79}                       \\
6exz              & 0.00              & \textbf{0.01}        &  \textbf{0.01}                                \\
6vw1              & 0.00           & 0.00                    & 0.00                                   \\
7mrx              & 0.00              & \textbf{0.54}        &  \textbf{0.54}                       \\ 
\midrule
pass rate         & 6/17           & \textbf{9/27}       &  \textbf{10/17}                              \\
avg. success rate & 0.06           & \textbf{0.26}        &  \textbf{0.29}                               \\ 
\bottomrule
\end{tabular}
}
\vspace{4mm}
\end{table}
\paragraph{Evaluation with more advanced folding model.}
Moreover, we also investigate evaluation with other structure prediction models, such as ESMFold~\cite{lin2022esmfold}. 
Results are shown in \Tabref{tab:motif_esmfold}, we consider the Alpha Carbon~(CA) pLDDT score predicted by ESMFold as the overall pLDDT score of the amino acid. 
We observe that ESMFold judges more strictly than OmegaFold. 
When we evaluate scaffold by ESMFold, there is a slight decline in the overall pass rate and average success rate, compared with the evaluation of OmegaFold.



\begin{table}[t!]
   \centering
   \small
   \setlength{\tabcolsep}{2pt}
   \vspace{-2.5mm}
   \caption{ Ablation study on the CATH4.3 benchmark, which w/ draft means that the reverse process is based on the $\xt{t}_{draft}$.}
   
   \label{tab:IF_ablation}

   \begin{tabular}{lrccc}
   \toprule
    \multirow{2}{*}{ Models} 
   & \multirow{2}{*}{ Trainable} 
   & \multirow{2}{*}{\metric{AAR}} 
   & \multicolumn{2}{c}{ struct. eval.} \\
    \cmidrule[0.3pt](lr){4-5} & Params. & & \metric{scTM}  & \metric{pLDDT} \\
   \midrule


   \textsc{LM-Design} (\textit{w/} draft)           &  6.3M/650M  & 56.49 &  0.85 & 74.89  \\
  \cmidrule[0.5pt](lr){1-5}
    \method                                                &  6.3M/650M   & 55.75   &  0.83  & 73.72  \\ 
   \chl \method (\textit{w/} draft)                                               & \chl 6.3M/650M   &\chl \textbf{56.61}   & \chl \textbf{0.86}  &\chl \textbf{76.78}  \\ 
  
\bottomrule
\end{tabular}
   \vspace{4mm}
\end{table}

\subsection{Structure-conditional generation: inverse folding}
\paragraph{Model architecture.}
\method only takes amino acid tokens as input, instead of structure formats such as 3D coordinates.
Therefore, in order to endow \method with structural awareness, we follow \textsc{LM-Design}~\citep{zheng2023structure} and place a structural adapter after the last layer of \method, which can attach the structure information to the original output probability. 
The overall architecture of the structural adapter is constituted by three components, \ie, a structure encoder, a \pLM as sequence decoder, and a structural adapter that bridges both. 
We can utilize an arbitrary pretrained structure encoder to process the 3D coordinates and provide structure information for \method.
For \pLMs as the sequence decoder side, we primarily used the \method, with its pretrained model weights. 
The structural adapter composes a multi-head attention that queries structure information from the structure encoder, followed by a \textit{bottleneck} feedforward network (FFN) to impose non-linearity and abstract features/representations. 
\textsc{RoPe}~\citep{su2021rope} was used the supplement multi-head attention for better modeling of positional information.
In all our experiments, only one structural adapter was placed after the last layer of \method, following~\citet{zheng2023structure}.

\paragraph{Training and inference details.}
During training, we freeze the parameters of the structure encoder and \method, only optimizing the structural adapter with the simplified discrete diffusion objective~\citep{zheng2023reparameterized}. 
However, we find that there is an exposure bias problem here: \method learns the reverse denoising process based on the ground truth context, \ie $\xt{t}$, which is obtained by adding noise on the ground truth sequence, \ie $\xt{0}$.
During inference, \method has to denoise given the context predicted, which is not always right, leading to training-inference inconsistency.
Therefore, we slightly modify the training objective in \Eqref{eq:reparam_obj}.
Specifically, we obtain $\xt{t}$ by adding noise on the draft sequence generated by the pretrained structure encoder, rather than the ground truth $\xt{0}$, which we refer to as $\xt{t}_{draft}$.
Then \method will learn the reverse process that reconstructs the $\xt{0}$ given the $\xt{t}_{draft}$, as shown in \Eqref{eq:IF_obj}.
Since the draft sequence is available both in training and inference time, the issue of exposure bias is mitigated.
We find this technique can further boost the performance of \method in the inverse folding task, as illustrated in the Tab.~\ref{tab:IF_ablation}

\begin{align}
\mathcal{J}_t 
 = \mathbb{E}_{q(\xt{0})} \bigg[\lambda^{(t)}  \sum_{1 \leq i \leq L} b_i(t) \cdot \log p_{\theta}(\xt{0}_i|\xt{t}_{\textrm{draft}})\bigg], 
\label{eq:IF_obj}
\end{align}

At inference time, we follow the \method generative process, except that we obtain protein sequence via greedy deterministic decoding, instead of random sampling from the distribution.
Besides, considering that we have had an unconditional model, i.e. the \method itself, and a conditional model, \ie, the \method with structural adapter, we can also seamlessly utilize the classifier-free guidance paradigm during inference.


\subsection{Classifier-free guidance}
\label{sec:classifier-free}
Classifier-free guidance~\citep{ho2021classifierfree} has been shown as an effective way to enhance conditional diffusion models.
Likewise, for \method, we can derive an implicit classifier using the Bayes rule
\begin{align}
q(\bm{y} | \xt{t-1}) & = q(\bm{y} | \xt{t-1}, \xt{t}) \nonumber \\
& = \frac{q( \xt{t-1} | \xt{t}, \bm{y}) }{q(\xt{t-1} | \xt{t})}q(\bm{y}|\xt{t}). \nonumber
\end{align}
If we already have an unconditional model $p_\theta(\xt{t-1}| \xt{t})$ and a conditional model $p_\theta(\xt{t-1} | \xt{t}, \bm{y})$ as the estimates, then by substituting this implicit classifier into Eq.~\ref{eq:classifier-guidance}, we can obtain
\begin{align}
    \xt{t-1} & \sim p_\theta(\xt{t-1}| \xt{t}) p_\phi(\bm{y} | \xt{t-1})^{\eta} \nonumber \\
    & \propto p_\theta(\xt{t-1} | \xt{t}) \big( \frac{p_\theta( \xt{t-1} | \xt{t}, \bm{y})}{p_\theta(\xt{t-1} | \xt{t})}\big)^{\eta} \nonumber \\
    & =  p_\theta( \xt{t-1} | \xt{t}, \bm{y})^{\eta} \cdot  p_\theta(\xt{t-1}|\xt{t})^{(1-\eta)},\nonumber
\end{align}
wherein when $\eta = 1$, it is equivalent to sampling from the original conditional \method without guidance, whereas $\eta > 1$, we not only prioritize the conditional model to contribute more but also discourage the samples from moving away from the unconditional distribution. 
In other words, it reduces the chance of generating samples that do not use conditioning information, in favor of the samples that explicitly do.

\begin{figure}[h!]
    \centering
    \vspace{2mm}
    \includegraphics[width=0.9\linewidth]{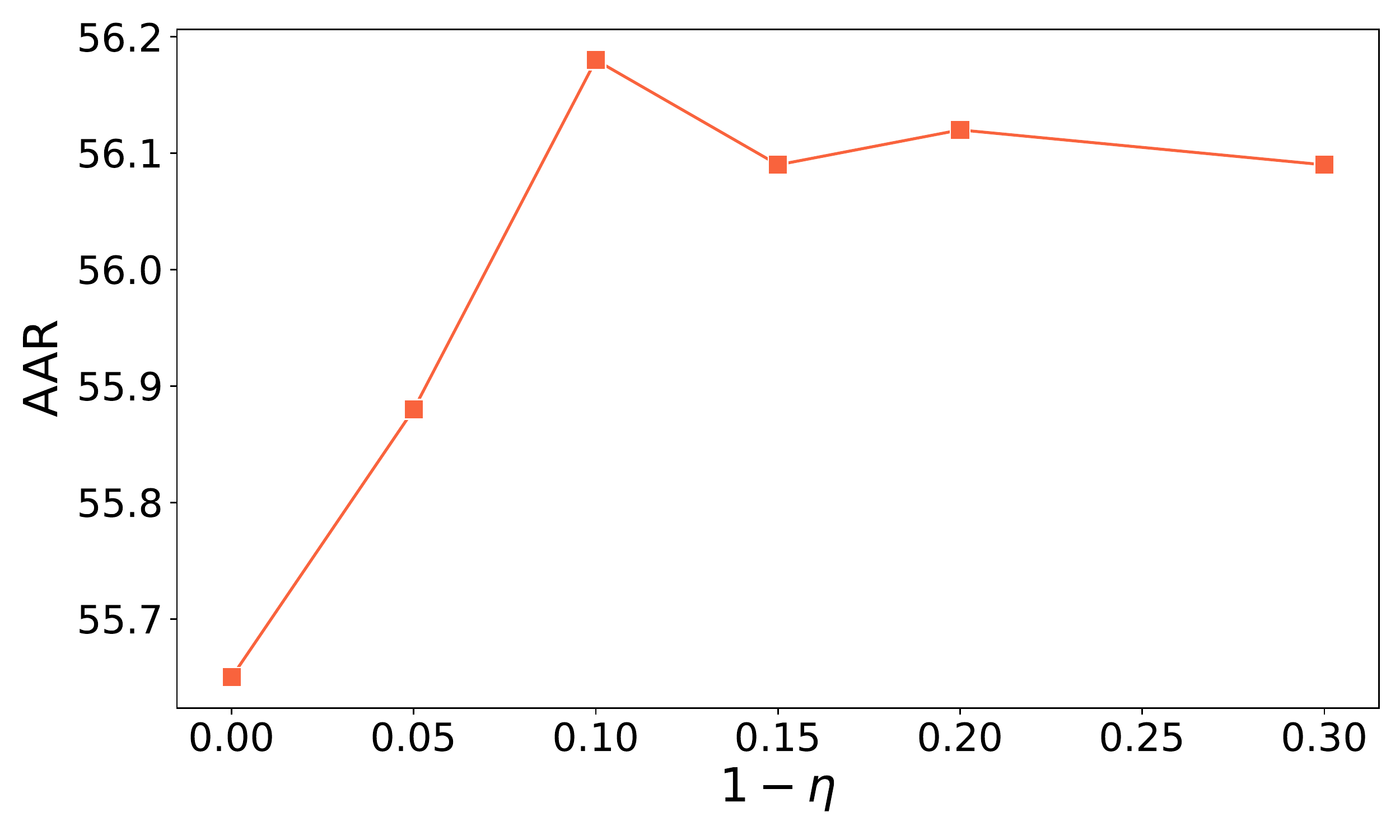}
    \vspace{-4mm}
    \caption{Classifier-free guidance enhances structure-conditioned sequence generation (inverse folding).} 
    \label{fig:classifier_free}
    \vspace{4mm} 
\end{figure}

Note that when we use adapter tuning to adapt \method for conditional generation, we only finetune the newly-added parameters, which means that we can already access both the unconditional model (original \method) and conditional model (the adapter-tuned model) simultaneously for free. 
As demonstrated in \Figref{fig:classifier_free} on structure-conditioned sequence generation, we can find that \method as a diffusion model can benefit from classifier-free guidance, improving its conditional generation immediately. 

\section{Related Work}
\label{sec:related}

\subsection{Language Models}
The dominant paradigm of language models is autoregressive language models, which breaks down the mutual distribution over the tokens of a sequence into conditional probabilities via the chain rule ${p(\bm{x}^{[1:N]})=\prod_{i=1}^N p(x^{[i]}|\bm{x}^{[1:i-1]})}$ and generates tokens by ancestral sampling from left to right~\citep{bengio2000neural,sutskever2014sequence,vaswani2017attention}.
Recently, researchers propose the non-autoregressive language models as an alternative~\citep{gu2018non}. 
These models do not need to obey the left to right generation order~\citep{qian2022diff,huang2022PDAT} and demonstrate competitive or superior performance compared to their autoregressive counterpart across a wide range of domains including languages~\citep{qian2021volctrans,huang2022PDAT,qian2022diff,huang2022PDAT,zheng2023reparameterized}, speeches~\citep{kim2021conditional}, proteins~\citep{zheng2023structure}, and molecules~\citep{hoogeboom2022equivariant}.
Among the numerous non-autoregressive language models, diffusion language models~\citep{li2022diffusionlm,gong2022diffuseq,zheng2023reparameterized} have emerged as a solid and promising framework.
Pretraining language models on a massive scale of unlabeled data markedly improves their downstream task performance~\citep{word2vec,elmo,radford2018gpt,devlin2019bert}.
As data volume and model sizes scale up, the training loss of language models predictably declines~\citep{kaplan2020scaling,hoffmann2022training,muennighoff2023scaling}, and enhancing downstream task performance even without specific tuning~\citep{gpt2}.
GPT3~\citep{brown2020gpt3} is a significant point in the journey, taking model sizes to 175B parameters, proposing in-context learning to bolster language models' competence in solving certain tasks with only a handful of demonstrations.
Furthermore, \citet{wei2021flanv1,t0,ouyang2022instructgpt} introduce instruction tuning, finetuning pretrained language models on series of tasks described via instructions, which elicits the instruction following ability of models and significantly enhances their zero-shot performance on unseen tasks.
More impressively, sufficiently large language models exhibit the emergent abilities such as multi-step reasoning~\citep{kojima2022zeroshotcot,wei2022emergent,wei2022CoT}, which small models do not possess~\citep{fu2023specializing}.
Empowered by large language models, helpful applications such as conversational AI systems\footnote{\url{https://chat.openai.com/}} and autonomous agents\footnote{\url{https://github.com/Significant-Gravitas/Auto-GPT}} have garnered much interest.
Although the most capable models at the moment are restricted in access, open-sourced efforts~\citep{zeng2022glm,touvron2023llama,touvron2023llama2,alpaca,vicuna2023,moss} have largely enhanced the public accessibility of powerful large language models.

\subsection{Protein Language Models}
Thanks for the abundance of 1D amino acid sequences, there is growing interest in developing protein LMs at the scale of evolution, such as the series of ESM~\citep{rives2019esm,lin2022esmfold}, TAPE~\citep{rao2019evaluating}, 
ProtTrans~\citep{elnaggar2021prottrans}, PRoBERTa~\citep{nambiar2020transforming}, PMLM~\citep{he2021pre}, 
ProteinLM~\citep{xiao2021modeling}, 
PLUS~\citep{min2021pre}, 
Adversarial Masked LMs~\citep{mcdermott2021adversarial}, 
ProteinBERT~\citep{brandes2022proteinbert}, 
CARP~\citep{yang2022convolutions} in masked language modeling (MLM) paradigm, 
ProtGPT2~\citep{ferruz2022protgpt2} in causal language modeling paradigm, and several others~\citep{melnyk2022reprogramming,madani2021deep,unsal2022learning, nourani2021tripletprot, lu2020self, sturmfels2020profile, strodthoff2020udsmprot}.
These protein language models exhibit remarkable generalization ability on various downstream tasks and be able to capture evolutionary information about secondary and tertiary structures from sequences alone.
Meanwhile, recent study shows these models' potency in revealing protein structures~\citep{lin2022esmfold}, predicting the effect of sequence variation on function~\citep{meier2021language}, antibody infilling~\citep{melnyk2022reprogramming} and many other general purposes~\citep{rives2019esm}.
Simultaneously, \citet{verkuil2022language} demonstrate that the large scale protein LMs can generate \textit{de novo} proteins by generalizing beyond natural proteins, both theoretically and experimentally validating their hypothesis in exhaustive detail, in which \pLMs demonstrate competency in designing protein structure despite being exclusively trained on sequences.



\subsection{Diffusion Language Models}
Derived from diffusion models~\citep{sohl2015diffusion}, diffusion language models is a variety of generative model that samples data via an iterative denoising process from noise. They can be divided into continuous~\citep{ho2020ddpm,song2020sde} and discrete~\citep{hoogeboom2021argmax,austin2021structured} categories according to the distribution they model.
Continuous diffusion models make great success in vision~\citep{dhariwal2021diffusionbeatgans,rombach2021highresolution,ho2022imagenvideo}, but they struggle in languages for operating on continuous surrogates of discrete tokens~\citep{li2022diffusionlm,gong2022diffuseq,han2022ssd,cdcd,yuan2022seqdiffuseq,gao2022difformer,ye2023dinoiser,chen2023cheaper,wu2023ar}, which has difficulty bypassing the pitfall of discreteness~\citep{ye2023dinoiser} and still lags behind autoregressive language models.
In contrast, discrete diffusion models, albeit having limited progress in large-scale applications, are innately suited to the data type inherent to languages (\ie, sequences of discrete tokens).
\citet{zheng2023reparameterized} makes commendable strides in discrete diffusion models and enhancing these models to yield comparable performance with autoregressive models on typical language generation benchmarks like machine translation.
Furthermore, as shown by \citet{he2022diffusionbert,zheng2023structure}, there are close relationship between discrete diffusion models and masked language models~(MLM), a widely adopted pretraining paradigm in NLP~\citep{devlin2019bert,liu2019roberta}.
Following this line, \citet{ye2023diffusion} propose scaling discrete diffusion LMs with diffusive adaptation, showing strong performance on several conditional text generation tasks, and accessing zero-shot instruction following, few-shot in-context learning and the promise of structured reasoning with instruction tuning.

\subsection{Protein Structure Diffusion Models}
Diffusion models have become popular tools in structural biology for protein generation, and their utility has been demonstrated across a range of generative tasks in recent years. \citet{trippe2022diffusion}, along with others, have introduced several diffusion model variants, each with its unique approach. For instance, while some models focus on generating the protein backbone by diffusing over protein coordinates, others, such as those proposed by \citet{wu2022high}, target inter-residue angles. \citet{lin2023generating} and \citet{yim2023framediff} have developed models that handle both the position and orientation of residue frames.
RFDiffusion~\citep{watson2023RFdiffusion} is a model that assists in designing protein structures for specific functions, such as enzymes. It is versatile in protein design and has been used to create therapeutic proteins, with some designs being confirmed in the laboratory.
ProteinSGM~\citep{lee2022proteinsgm} is a model that uses 2D matrices, which represent the distances and angles between protein parts, to create 3D protein structures for novel protein designs.
FoldingDiff~\citep{wu2022protein} is a model that generates protein sequences expected to fold into a specific structure. These sequences are verified with prediction tools, although they have not been experimentally confirmed yet.
Chroma~\citep{ingraham2023chroma} is a model designed for creating large proteins and protein complexes, considering various constraints like distances and symmetry. It transforms a collapsed polymer into protein backbone and sequence more quickly than older methods, thereby allowing for the efficient generation of large structures.

\subsection{Protein Inverse Folding}
The structure-based protein sequence design is typically formulated as a conditional sequence generation problem by deep generative modeling, wherein protein 3D structures are usually depicted as a \textit{k}-NN graph~\citep{ingraham2019generative}.
The protein graph establishes edge features between adjacent residues and encodes residue information as node features, modeled by graph neural networks (GNNs).
GraphTrans~\citep{ingraham2019generative} and GVP~\citep{jing2020gvp} utilizes the graph attention encoder and autoregressive decoder for protein design.
Recently, ProteinMPNN~\citep{dauparas2022proteinmpnn} and PiFold~\citep{gao2022pifold} introduce more complex protein features and expressive GNNs, resulting in significant improvements. 
Furthermore, in addition to the primary generative purpose, this task can also be used as a proxy for protein (structure-aware) representation learning~\citep{yang2022masked}.
A critical and significant challenge herein is the lack of sufficient protein structure data.
To this end, ESM-IF~\citep{hsu2022esmif} alleviate this issue with effective data augmentation by back-translation with AlphaFold 2~\cite{jumper2021AF2}.  , resulting in dramatic improvements. 
On the other hand, \citet{zheng2023structure} demonstrate how to efficiently steering large pretrained protein LMs into a structure-informed sequence generative models in a mask-predict generative manner, attaining state-of-the-art results on single-chain and complex protein benchmark.
Most recently, graph diffusion models have also been studied for inverse folding problem~\citep{yi2023gradeif}.

\newpage
\clearpage

\onecolumn
\section{Visualization of Unconditional Samples}

\begin{figure*}[h!]
\vspace{10mm}
    \centering
    \includegraphics[width=\linewidth]{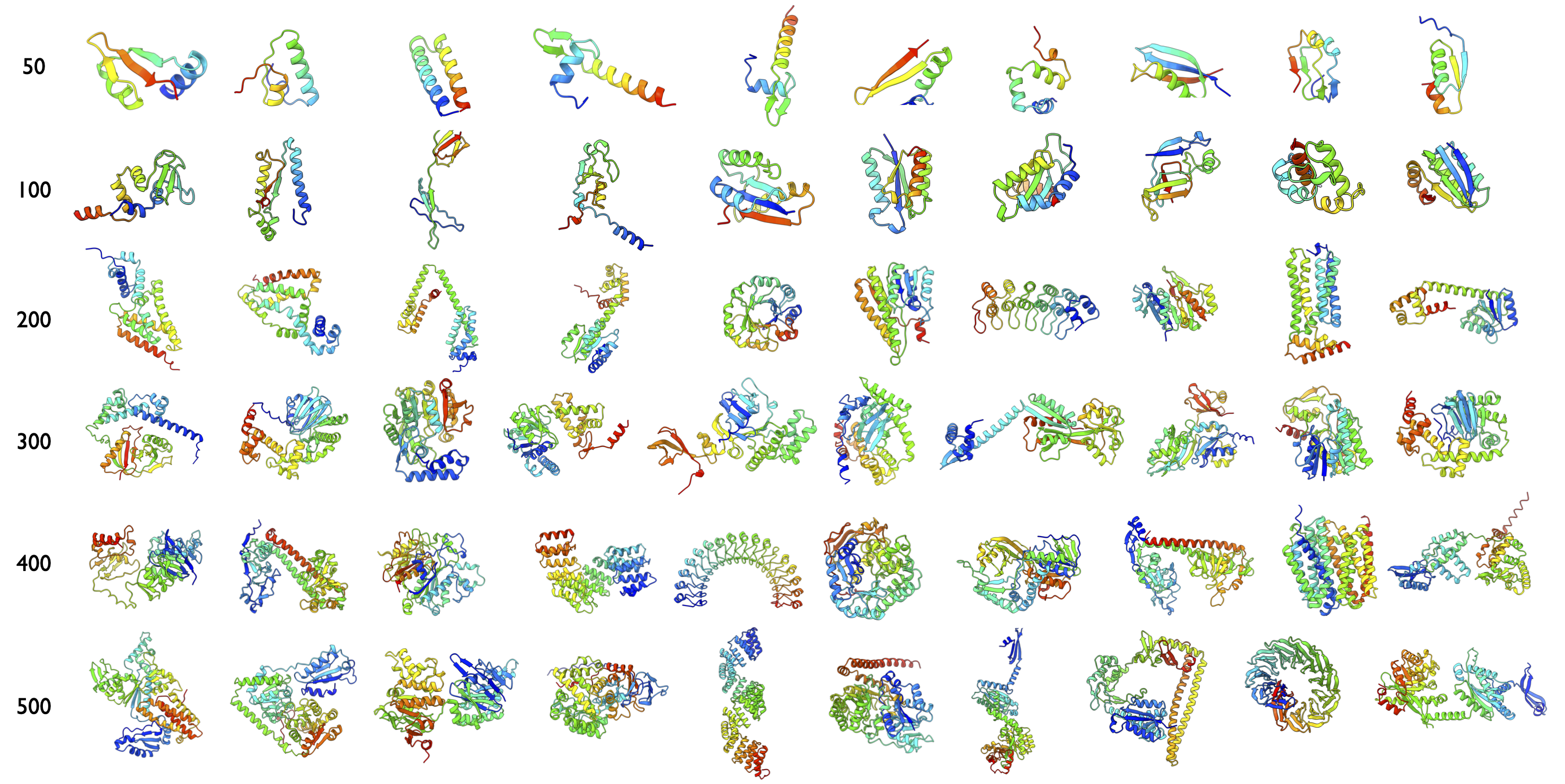}
    \vspace{-4mm}
    \caption{Visualized examples from 50 to 500 in length.} 
    \label{fig:big_pic_100_500}
    \vspace{20mm}
\end{figure*}

\par

\begin{figure*}[h!]
    \centering
    \includegraphics[width=\linewidth]{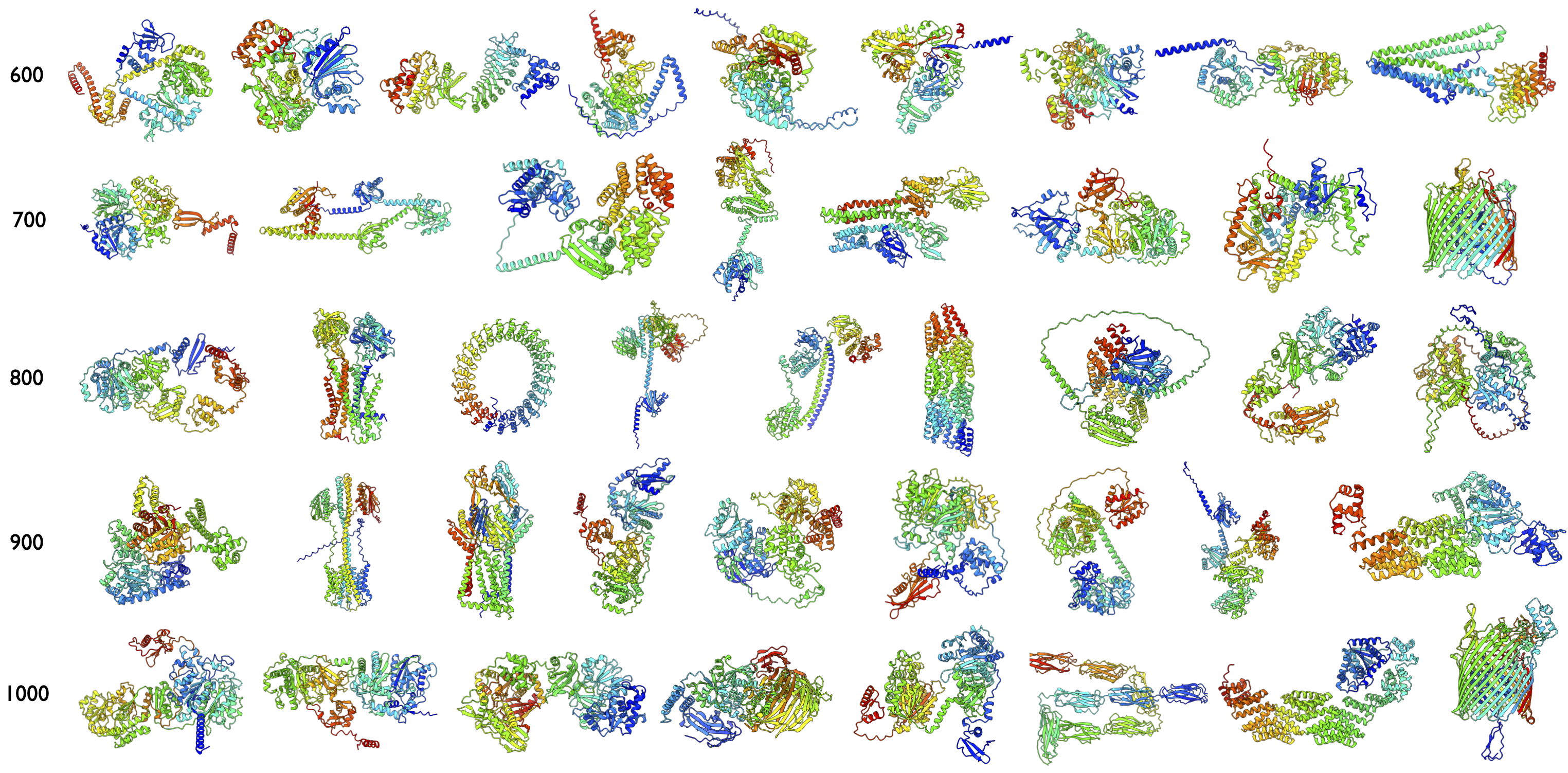}
    \vspace{-4mm}
    \caption{Visualized examples from 600 to 1000 in length.} 
    \label{fig:big_pic_600_1000}
\end{figure*}


\end{document}